\newif\ifacm
\acmtrue
\newif\ifcomment
\commenttrue
\newif\ifcameraready
\camerareadyfalse
\newif\ifwatermark
\watermarkfalse

\ifwatermark
      \usepackage{draftwatermark}
      \SetWatermarkScale{.7}
      \SetWatermarkText{\shortstack{In Submission:\\Do Not Distribute}}
      \SetWatermarkColor[rgb]{1,0.88,0.88}
\fi

\ifcomment
    \newcounter{MVNumberOfComments}
    \newcounter{YZNumberOfComments}
    \stepcounter{MVNumberOfComments}
    \stepcounter{YZNumberOfComments}
    \newcommand{\mvnote}[1]{\textcolor{blue}{\small \bf [MV\#\arabic{MVNumberOfComments}\stepcounter{MVNumberOfComments}: #1]}}

    \newcommand{\NOTE}[1]
    {
      {\footnotesize\it
        \begin{center}
          \begin{tabular}{|c|}
           \hline
            \parbox{0.85\columnwidth}{
              \medskip
              #1
              \medskip} \\
            \hline
          \end{tabular}
        \end{center}
        }
    }
\else
    \newcommand\mvnote[1]{}
    \newcommand\NOTE[1]{}
\fi

\newcommand{\eg}{{e.g.,}\xspace}
\newcommand{\ie}{{\it i.e.,}\xspace}

\newcounter{NumTakeaways}
\stepcounter{NumTakeaways}

\newcommand{\tool}{{WebDiffusion}\xspace}
\newcommand{\numpagesLarge}{{200}\xspace}
\newcommand{\numimagesLarge}{{1,870}\xspace}
\newcommand{\numpartLarge}{{1,000}\xspace}
\newcommand{\numpages}{{60}\xspace}
\newcommand{\numimages}{{409}\xspace}
\newcommand{\numpart}{{940}\xspace}
\newcommand{\tot}{{9,400}\xspace}
\newcommand{\numiter}{{20}\xspace}
\ifacm
    \documentclass[sigconf]{acmart}
\else 
    \documentclass[10pt, conference, letterpaper]{IEEEtran}
\fi

\ifacm
\setcopyright{none}
\usepackage{gensymb}
\usepackage{booktabs}  % for toprule, bottomrule in tables
\usepackage{pbox}
\usepackage{subfigure}
\usepackage{epsfig,endnotes}
\usepackage{epstopdf}
\usepackage{times}
\usepackage{color}
\usepackage{colortbl} % to allow background color in the table
\usepackage{graphicx}
\usepackage{xspace}
\usepackage{enumitem}
\usepackage{multirow}
\usepackage{tikz}
\usepackage[ruled,vlined,linesnumbered]{algorithm2e}
\usepackage{lmodern}
\usepackage[T1]{fontenc}
\usepackage[utf8]{inputenc}
 
\else
\usepackage{cite}
\usepackage{amsmath,amssymb,amsfonts}
\usepackage{algorithmic}
\usepackage{graphicx}
\usepackage{textcomp}
\usepackage{xcolor}
\def\BibTeX{{\rm B\kern-.05em{\sc i\kern-.025em b}\kern-.08em
    T\kern-.1667em\lower.7ex\hbox{E}\kern-.125emX}}

\usepackage{gensymb}
\usepackage{booktabs}  % for toprule, bottomrule in tables
\usepackage{pbox}
\usepackage{subfigure}
\usepackage{epsfig,endnotes}
\usepackage{epstopdf}
\usepackage{times}
\usepackage{color}
\usepackage{colortbl} % to allow background color in the table
\usepackage{graphicx}
\usepackage{xspace}
\usepackage{enumitem}
\usepackage{multirow}
\usepackage{tikz}
\usepackage[ruled,vlined,linesnumbered]{algorithm2e}
\usepackage{lmodern}
\usepackage[T1]{fontenc}
\usepackage[utf8]{inputenc}
\fi

\setcopyright{none}
\renewcommand\footnotetextcopyrightpermission[1]{} % removes footnote with conference information in first column
\begin{document}

% Title and authors go here
%\title{Exploring the Potential of Generative AI for the Evolution of the World Wide Web}
\title{Exploring the Potential of Generative AI for the World Wide Web}
\subtitle{\url{https://webdiffusion.ai/}}

\ifacm
% [SIGCONF] Title and authors go here
    \renewcommand{\shortauthors}{Zaki et al.}
    % \author{Paper \#1627}
    \author{Nouar AlDahoul$\star$, Joseph Hong$\star$, Matteo Varvello$\Delta$, Yasir Zaki$\star$}
    \affiliation{%
      \institution{
      $\star$~New York University Abu Dhabi, $\Delta$~Nokia~Bell Labs}
      \country{
        $\star$~United Arab Emirates,
        $\Delta$~United States of America
        }
    }
    \email{yasir.zaki@nyu.edu,  joseph.hong@nyu.edu, matteo.varvello@nokia.com}
\else
    \author{Paper \#XXX}
    % \author{
    %     \IEEEauthorblockN{1\textsuperscript{st} Given Name Surname}
    %     \IEEEauthorblockA{\textit{dept. name of organization (of Aff.)} \\
    %     \textit{name of organization (of Aff.)}\\
    %     City, Country \\
    %     email address or ORCID}
    %     \and
    %     \IEEEauthorblockN{2\textsuperscript{nd} Given Name Surname}
    %     \IEEEauthorblockA{\textit{dept. name of organization (of Aff.)} \\
    %     \textit{name of organization (of Aff.)}\\
    %     City, Country \\
    %     email address or ORCID}    
    % }
\fi 

% 
% \ifcameraready
%     \ifacm
%     \begin{ACMkeywords}
%             component, formatting, style, styling, insert
%     \end{ACMkeywords}
%     \else
%         \begin{IEEEkeywords}
%             component, formatting, style, styling, insert
%         \end{IEEEkeywords}
%     \fi
% \if

% add an abstract and title 
\ifacm
    \begin{abstract}
Generative Artificial Intelligence (AI) is a cutting-edge technology capable of producing text, images, and various media content leveraging generative models and user prompts. Between 2022 and 2023, generative AI surged in popularity with a plethora of applications spanning from AI-powered movies to chatbots. In this paper, we delve into the potential of generative AI within the realm of the World Wide Web, specifically focusing on image generation. Web developers already harness generative AI to help crafting text and images, while Web browsers might use it in the future to locally generate images for tasks like repairing broken webpages, conserving bandwidth, and enhancing privacy. To explore this research area, we have developed \textit{\tool}, a tool that allows to simulate a Web powered by stable diffusion, a popular text-to-image model, from both a client and server perspective. \tool further supports crowdsourcing of user opinions, which we use to evaluate the quality and accuracy of 409 AI-generated images sourced from 60 webpages. Our findings suggest that generative AI is already capable of producing pertinent and high-quality Web images, even without requiring Web designers to manually input prompts, just by leveraging contextual information available within the webpages. However, we acknowledge that direct in-browser image generation remains a challenge, as only highly powerful GPUs, such as the A40 and A100, can (partially) compete with classic image downloads. Nevertheless, this approach could be valuable for a subset of the images, for example when fixing broken webpages or handling highly private content.
\end{abstract}
    \maketitle
    \pagestyle{plain}
\else 
    \maketitle
    \begin{abstract}
Generative Artificial Intelligence (AI) is a cutting-edge technology capable of producing text, images, and various media content leveraging generative models and user prompts. Between 2022 and 2023, generative AI surged in popularity with a plethora of applications spanning from AI-powered movies to chatbots. In this paper, we delve into the potential of generative AI within the realm of the World Wide Web, specifically focusing on image generation. Web developers already harness generative AI to help crafting text and images, while Web browsers might use it in the future to locally generate images for tasks like repairing broken webpages, conserving bandwidth, and enhancing privacy. To explore this research area, we have developed \textit{\tool}, a tool that allows to simulate a Web powered by stable diffusion, a popular text-to-image model, from both a client and server perspective. \tool further supports crowdsourcing of user opinions, which we use to evaluate the quality and accuracy of 409 AI-generated images sourced from 60 webpages. Our findings suggest that generative AI is already capable of producing pertinent and high-quality Web images, even without requiring Web designers to manually input prompts, just by leveraging contextual information available within the webpages. However, we acknowledge that direct in-browser image generation remains a challenge, as only highly powerful GPUs, such as the A40 and A100, can (partially) compete with classic image downloads. Nevertheless, this approach could be valuable for a subset of the images, for example when fixing broken webpages or handling highly private content.
\end{abstract}
\fi 

% paper sections 
%%%%%%%%%%%%%%%%%%%%%%%
\section{Introduction}
\label{sec:intro}
%%%%%%%%%%%%%%%%%%%%%%%
In the last year, generative Artificial Intelligence (AI) has emerged as a revolutionary technology capable of generating diverse forms of content, including text~\cite{aaa,bbb}, images~\cite{Ramesh_2021_DALLE, Rombach_2022_CVPR, midjourney, Nichol_2021_GLIDETP,Ramesh_2022_DALLE2}, and multimedia~\cite{kumar2023comprehensive,ampermusic,plaugic_2017}. This surge in interest has ignited a wave of innovative applications across various domains, from the entertainment industry with AI-powered movies~\cite{zhu2023moviefactory, singer2022makeavideo} to the realms of medicine~\cite{ccc,mesko2023imperative}. In this paper, we delve into the fascinating role of generative AI within the context of the World Wide Web. 

Web developers are turning to generative AI to automate the creation of images and multimedia assets, or speedup webpages.\footnote{According to HubSpot Blogs research, 93\% of web designers have used an AI tool to assist with a web design-related task in the summer of 2023~\cite{ai_web_usage}.} Similarly, browser vendors  are exploring the potential of generative AI running inside the browser, \eg Summarizer by Brave~\cite{brave_summarizer}  and MAX by Arc~\cite{arc_max}. In the future, Web media like images could be locally generated by the browser offering solutions to rectify broken web pages, optimize bandwidth usage, and enhance privacy protection. 

To explore this intriguing intersection between generative AI and the Web, we introduce \tool, a tool that allows to emulate a Web leveraging stable diffusion for image generation, both from a client and a server perspective. We focus on image generation for two reasons. First, images are a vital component of the Web: for example, the median webpage today is comprised of 900 KB of images (or about 44\% of its total size)~\cite{kelton2021browselite}. Second, image generation has no impact on Webcompat~\cite{jscleaner,webmedic, muzeel}, differently from AI-based generation of HTML, JavaScript, etc. Still, this is an interesting research question which we left open for future work. 

Given a target webpage, \tool crawls its content, produces AI-based images, and performs experiments with unmodified browsers. Last but not least, \tool integrates with Prolific~\cite{prolific}, a popular crowdsourcing platform, to gather worldwide eyeballs reporting on the quality of Web images and webpages produced by generative AI. This component of \tool is currently live at \textit{https://webdiffusion.ai/}. We use \tool to investigate several research questions. 

\begin{list}{$\bullet$}{\leftmargin=0.5em \itemindent=0em}
    % \item \textbf{Can Web images be automatically generated?} 
    %We crawled the top 500 webpages from Trancos’ top million list~\cite{tranco}. After filtering\yznote{We might get asked what kind of filtering? I think we need to come up with a better reasonning here}, we end up with \numpagesLarge webpages from which we collected and automatically annotated\yznote{Not sure what it means here by automatically annotated? Is this referring to the text gathering where the image is present in the page? I don't think we did image-to-text here} \numimagesLarge images. Next, we crowdsourced \numpartLarge people to evaluate the quality of these annotations, estimating a success rate of 80\% for the client-side approach, and close to 90\% for the server-side approach. \mvnote{client versus server is not currently defined -- make a decision }\yznote{I think this ties back to my previous comment. It is not clear what annotation means here. We might want to explain this a bit above.}
    \item \textbf{Can Web images be automatically generated?} We identify the top 200 webpages from Tranco's top one million which contain at least an image, load correctly, and are written in English. From these pages, we collect 1,870 images for which we generate annotations (prompts) to be used for image generation. We consider client-side information only, \ie alt-text and the content of the webpage, to emulate a generative AI running in the browser, and server-side information, \ie with the addition of the actual image description and details as an approximation of the Web designer intent. With a user study involving 1,000 participants, we show that the client-based approach produces relevant annotations 80\% of the time, while this fraction grows to 90\% when considering a server-based approach. 

    \item \textbf{What is the quality of AI-generated webpages?} From the above dataset, we randomly select \numpages webpages (accounting for \numimages unique images) and generate each image (and webpage) using stable diffusion. To do so, we rely on the annotations discussed above, thus producing both client-based and server-based versions of each image/webpage. We then crowdsource about \numpart people (10 per image/webpage, on average) to score the quality of the AI-generated content. We find that 70-95\% of the images are scored as either ``fair'' or higher (``good'', ``very good'', or ``excellent''), and 95\% of webpages are scored as ``good'' or higher. The server-based approach achieves the highest quality, thanks to the additional contextual information available. We further investigate which image features, \eg landscape versus a celebrity, have the largest impact on high and low scores. We find that images including text achieve the lowest scores, followed by images with faces.
        
    %This is a challenging research question as AI generated images are not expected to be equivalent to an input image, especially when such image does not exist. In the context of the Web, however, we have a large corpus of webpages for which images have been already selected, making it the perfect data-set to study how well text-to-image generation currently works. 

    \item \textbf{Can image generation run in the client?} By loading multiple webpages under several network conditions, our experiments show that, as of today, only powerful GPUs (like Nvidia Tesla A40 and A100) can compete with actual image download, \eg not dramatically slowdown a webpage load. Further, this is only true for \texttt{SpeedIndex}, a popular web performance metric relating to only the visible portion of the screen (i.e., what is known as above the fold). We found that there is potentially a 2.5-5 seconds improvement for \texttt{SpeedIndex} leveraging image generation. As we move to \texttt{PageLoadTime}, which instead requires the whole page to be downloaded, then even powerful GPUs would add tens of seconds. We thus conclude that such approach is currently viable for applications operating on a handful of images, such as privacy-preserving advertising and webpage repairing. 
\end{list}

%%%%%%%%%%%%%%%%%%%%%%%%%%%%%%%%%%%%%%%%%%%%
\section{Background and Related Work}
\label{sec:related}
%%%%%%%%%%%%%%%%%%%%%%%%%%%%%%%%%%%%%%%%%%%%
\noindent
\textbf{Text-to-Image Generation} -- A variety of text-to-image generation models  have been released over the last few years. These models face the so-called \textit{generative learning trilemma}~\cite{Denoising_Diffusion}, \ie they can only satisfy two out of three requirements that real-world applications of generative models should possess: high-quality outputs, output sample diversity, and computationally fast and inexpensive sampling. Image generation usually requires high-quality outputs, thus general adversarial network (GAN) or diffusion models are preferred.

Thanks to recent developments, diffusion models are currently outperforming GAN models~\cite{guided_diffusion_outperform_GANs}. These gradually add Gaussian noise to the input, and afterwards perform a reverse denoising process, which reverses this process and outputs an image~\cite{Denoising_Diffusion}. GLIDE and DALL-E 2~\cite{Nichol_2021_GLIDETP, Ramesh_2022_DALLE2} both use classifier-free diffusion models as a part of their image generation process. We found that these diffusion models produce very realistic outputs and have large mode coverage, ideal for image generation, but tend to be very slow due to the number of iterations required throughout the denoising processes.

This is the issue that the latent diffusion model~\cite{Rombach_2022_CVPR}, the foundation of \textit{stable diffusion}~\cite{stable}, addresses. By transforming the diffusion model to be compatible with a compressed image representation in a latent space, latent diffusion allows the retention of high-quality outputs on a wide variety of image-generation tasks quickly and in a computationally efficient manner~\cite{stable}. Fulfilling the three aforementioned requirements, latent diffusion has potential for real-world applications, which is our motivation for selecting stable diffusion as the image generation model for this paper. 

\vspace{0.05in}
\noindent
\textbf{Image Generation for the Web} -- To the best of our knowledge, no previous work has investigated the role of generative AI for Web images consumption. The closest work to this topic is~\cite{wang2021web}, where the authors propose to replace Web images with \textit{similar} content, \eg a similar picture of the Golden Gate bridge taken from a different angle. Their rationale is to ``race'' similar Web images -- identified via reverse image searches -- to speedup webpage loads. The similarity with our work stands in the idea that Web images are not always ``strict'', aka users might be willing to accept similar images to the originals if this comes with some benefits, such as page speedups in~\cite{wang2021web}. The key difference with our work is that we broadly explore the field of generative AI when applied to Web images, considering both a client and server standpoint.  
\begin{figure}[t]
    \includegraphics[width=3in]{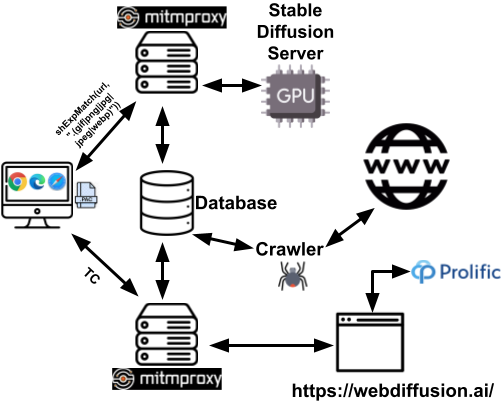}
    \caption{A visualization ot \tool, our tool to evaluate the applicability of text-to-image generation for the Web.}
    \label{fig:setup}
\end{figure}

%%%%%%%%%%%%%%%%% 
\section{\tool}
\label{sec:meth}
%%%%%%%%%%%%%%%%% 
%As discussed in Section~\ref{sec:intro}, generative AI is already used by Web designers, \eg to write articles and choose images, and is starting to appear inside browsers~\cite{brave_summarizer, arc_max}. 
This section presents \texttt{\tool},  a tool which allows to explore and emulate a Web leveraging stable diffusion for image generation. \tool aims at evaluating both a \textit{server} and a \textit{client} mode. The server mode refers to a scenario where Web designers rely on stable diffusion for image generation. The client mode refers to a scenario where image generation is fully run in the browser, \eg to generate privacy-preserving advertisement or fix broken URLs.  

At high level, \tool consists of the four components visualized in Figure~\ref{fig:setup}. First, a \textit{crawler} which makes local copies of actual webpages, while annotating embedded images with textual prompts using two techniques discussed next. Second, a \textit{proxy} system to replay webpages while replacing images with textual prompts to feed to a \textit{stable diffusion server}, allowing the emulation of an AI-powered Web without requiring an in-browser implementation. Finally, a \textit{Web interface} which collects human feedback on the accuracy of AI-generated images and webpages. In the remainder of this section, we describe each component of \tool.

%A set of technical solutions which allow to take advantage of AI-based text-to-image generation for the Web. We first overview three solutions to \textit{annotate} images, \ie generate image description which should be used by the browser for image generation. Next, we present several candidate approaches for text-to-image generation and why we converged to stable diffusion~\cite{stable_diff}.

\begin{figure}[t]
    \includegraphics[width=3in]{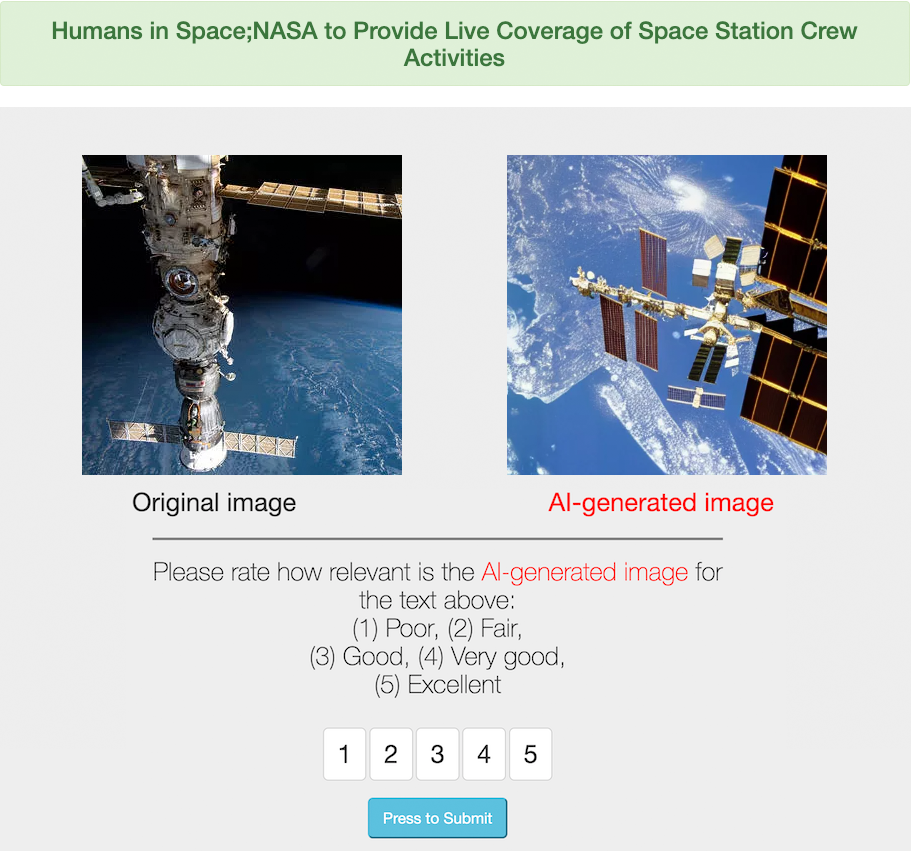}
    \caption{\tool's GUI. The top of the figure shows the text originated using the client-based approach, \ie eventual \textit{alt text} and nested divs information. The center of the figure shows both the original and the AI-generated images. The bottom of the figure allows study participants to score the quality of this AI-generated image, between 1 (poor) and 5 (excellent).}
    \label{fig:user_study}
\end{figure}

%%%%%%%%%%%%%%%%%%%%%
\subsection{Crawler}
\label{sec:meth:anno}
%%%%%%%%%%%%%%%%%%%%%%
The first task of the \tool's crawler is to generate local copies of webpages. These webpages can be analyzed to investigate the feasibility of image annotation techniques discussed below, as well as ``replayed'' to benchmark the performance of AI-generated webpages. The crawler consists of a Selenium~\cite{selenium} application which, for a given webpage, records headers and data returned for each HTTP(S) request along with its duration. Crawled webpages are stored in a database so that they can then be \textit{replayed} via the proxy. % while emulating their original duration. 

Next, the crawler runs the following annotation schemes to obtain textual prompts for each image embedded in a given webpage. Each textual prompt is also stored in the database along with the technique used for its derivation. %In the following, we describe the two techniques we have devised. 

\vspace{0.05in}
\noindent
\textbf{Client-Based}: This technique aims at being deployed today without any server-side support, meaning that it works solely relying on \textit{contextual} information, \eg \texttt{alt text} sometimes provided along with an image (about 50\% of the time~\cite{webalmanac2022}). The trade-off is a potential lack of accuracy as the browser is forced to ``guess'' the content of an image without having access to it. 

This mechanism works as follows. For each image tag in a page's HTML that has either a \texttt{src} or \texttt{background-image: url} attribute in its CSS styling, we look for its \texttt{alt text} (or \texttt{alt tag}) whose goal is to provide a description of an image in case the actual image is missing. In addition, we also iterate through the different \texttt{div} elements associated with the image looking for potential text within $<p>$, $<h1>$, $<h2>$, $<h3>$ and $<h4>$ tag elements, and then iterate through their children nodes to extract the text associated with them. This is often where the title and the description of an image are stored. We accumulate all the text associated with these different elements into forming the client-based textual prompt to be used to generate the image. It also recursively traverses the parent div nodes of the image div to extract all the text related to the image; it stops when it discovers the presence of another image div and reverts back to previous node.

\vspace{0.05in}
\noindent
\textbf{Server-Based}: This technique implies some server side support, \eg from Web developers, to annotate images. It extends the client-based approach with image annotation obtained via Microsoft's \texttt{GenerativeImage2Text} transformer~\cite{33,34}, which derives a textual prompt related to an input image. Note that leveraging \texttt{GenerativeImage2Text} allows us to build an approximation which can be evaluated today without support from Web developers.
%While in the future Web developers might directly annotate their images, 

%%%%%%%%%%%%%%%%%%%%%%%%%%%
\subsection{Proxy System}
%%%%%%%%%%%%%%%%%%%%%%%%%%%
In theory, stable diffusion can be integrated in any open source Web browser. In practice, this is a time consuming effort which can anyway lead to suboptimal performance as browsers are very complex software. Instead, \tool offloads image generation to an external machine (stable diffusion server) and leverages a proxy system to intercept and operate on Web traffic (see Figure~\ref{fig:setup}). This setup allows \tool to control how images should be served, either in a classic way or generated by the stable diffusion server. 

As shown in Figure~\ref{fig:setup}, legacy clients -- mobile or desktop running any browser -- are instructed to forward all their HTTP(S) traffic via  \texttt{mitmproxy}~\cite{mitmproxy}. This requires installing mitmproxy's root CA (Certificate Authority) to handle HTTPS traffic. Note that this step is only required for testing purposes, as a browser implementation would not require these steps. We assume all traffic is HTTP/2~\cite{h2RFC}. 

Our experimental proxy system consists of two \texttt{mitmproxy} proxies: one handling only requests for images, and one serving all the other requests. These two proxies are set in the testing client using a simple Proxy Auto-Configuration (PAC) file~\cite{pac_file} with a regular expression matching requests for images towards one proxy and vice-versa.\footnote{Note that this also requires to set the browser's   \texttt{PacHttpsUrlStrippingEnabled} policy to false so as to enable full URL visibility in the PAC file with HTTPS.} This is necessary to allow the flexibility to emulate network connectivity (realized via Linux \texttt{tc}~\cite{tc}) at the link between legacy clients and the proxy not handling image requests. In fact, the network characteristics of this link should not affect the images generated by the stable diffusion server, since we are emulating a client-side implementation. This is ensured as long as a fast connection (1~Gbps) is available between client, image proxy, and stable diffusion server. 
% image proxy and the client, or by colocating the image proxies with the same machine where the client runs. 

%and some potential network latency to emulate. After the latency expires, the content is returned to the client. \mvnote{@yasir -- mmm this kinda fight with Linux tc no?}\yznote{It does. So I don't know, shall we remove it?}  
For each request received, our proxy system performs a database lookup. For non-images, this lookup returns the content to be served. For images, the database lookup might return some textual prompt for image generation. This depends on the experiment running, \eg some prompts might be missing when testing the client-based approach (see Section~\ref{sec:meth}). In the presence of a textual prompt, the proxy contacts the stable diffusion server which generates the image. The stable diffusion server can be configured with different hardware to emulate different client capabilities. When the image is generated, it is returned to the proxy which forwards it back to the client. 

%%%%%%%%%%%%%%%%%%%%%%%%%%%%%%%%%%%%%
\subsection{Stable Diffusion Server}
\label{sec:meth:stable}
%%%%%%%%%%%%%%%%%%%%%%%%%%%%%%%%%%%%%
Given the prompts derived using any of the above techniques, we then rely on stable diffusion for automated image generation. We chose stable diffusion for the reason discussed in Section~\ref{sec:related}, and because of its open source nature. We deployed the stable diffusion (SDXL) implementation of \texttt{huggingface.co} using their \texttt{stabilityai/stable-diffusion-xl-base-1.0} \\model~\cite{31}\cite{32}. It is a latent diffusion model with a three times larger UNet backbone for text-to-image synthesis that shows remarkable performance compared to previous versions of stable diffusion. SDXL uses a second text encoder OpenCLIP ViT-bigG/14 and multiple novel conditioning schemes~\cite{32}. To create prompts from images, as needed by the server-based technique explained earlier, we have also utilized the large-sized \texttt{GenerativeImage2Text} transformer from \texttt{microsoft/git-large-coco}~\cite{33}\cite{34} which was fine-tuned on COCO to generate an image caption to be used as a text prompt. The architecture of the \texttt{GenerativeImage2Text} transformer consists of one image encoder and one text decoder under a single language modeling task~\cite{33}.

The \textit{quality} of images produced by stable diffusion increases with the number of inference steps, which in turn requires a longer generation time. The number of inference steps relates to how many denoising steps the model will use to improve the quality of the generated images. Generally, 50 denoising steps are sufficient to generate high-quality images. While experimenting with stable diffusion in the context of the Web, we realized that some textual prompts benefit more than others from more iterations, \eg human faces versus a landscape. In this paper, we settle for a constant number of iterations (\numiter) for fairness between images from different webpages. This number was chosen as a good compromise between quality and speed (about 1 second image generation time on our machine). From the different Web images that were generated, we have noticed that the above number of inference is enough for generating a large portion of the Web images, where any image that does not contain text, or faces had a good quality (for example food, landscape, objects, animals, etc.). However, an interesting direction for future work is to explore some dynamic stable diffusion settings based on the content of a textual prompt, and potentially additional webpage information. %We refer the reader to Section~\ref{sec:disc} for more details. 
Apart from the number of inference, all other stable diffusion parameters were kept to the default values: guidance scale (5), and width and height (1024x1024 pixels). We randomly generated the seed to have a different images generated each time.

Our stable diffusion server currently runs on a Linux server equipped with Nvidia Tesla A40 GPU~\cite{gpu} with a 48 GB GDDR6 with error-correcting code (ECC), 300 W maximum power consumption, and a 696 GB/s GPU memory bandwidth. The server specifications are: Intel@Xeon(R) Bronze 3204 CPU@1.90GHz x 12, with 16 GiB RAM, and running Ubuntu 20.04.5 LTS.

%%%%%%%%%%%%%%%%%%%%%%%%%%%
\subsection{Web Interface}
\label{sec:meth:gui}
%%%%%%%%%%%%%%%%%%%%%%%%%%%
Last but not least, \tool offers a Web interface at \textit{https://webdiffusion.ai/}. At this page, users can explore Web images or full webpages automatically generated using the above annotation schemes (see Appendix Figures~\ref{fig:nasa},~\ref{fig:etsy} and~\ref{fig:cnn}), while providing some feedback on the quality of the generated content. The Web interface accepts a parameter (\texttt{?type=[images, webpages]}) to control whether to show images or full webpages, and which annotation scheme to use -- default is server-based, while client-based is activated by adding the suffix \texttt{\_client}.

Figure~\ref{fig:user_study} shows a visualization of \tool's Web interface when instrumented to show a image comparison. At the top of the page some textual prompt is offered to describe the images. This text was obtained using the techniques described above for image annotation. Next, two images are shown. One image is an original Web image pulled from a given website; the other image is originated via stable diffusion using the above textual prompt. Original images are displayed on the left, whereas the AI-powered images are displayed on the right. Both images are clearly labeled. At the bottom of the page, the user is asked to rank the quality of AI-generated images between 1 (poor) and 5 (excellent). We integrated \tool's Web GUI with Prolific~\cite{prolific}, a popular  crowdsourcing website which allows us to quickly scale user studies. We used a standard 5-point Likert scale with a theme focusing on evaluating quality. The theme uses the following labels (also known as ``Likert Scale Response Anchors''): (1) poor, (2) fair, (3) good, (4) very good, and (5) excellent. Notice that we have also provided the numerical values associated with each label so as to give the participants a sense of the distance between the labels, and to help use process the data qualitatively during our analysis phase.
\begin{figure}[t]
    \includegraphics[width=3in]{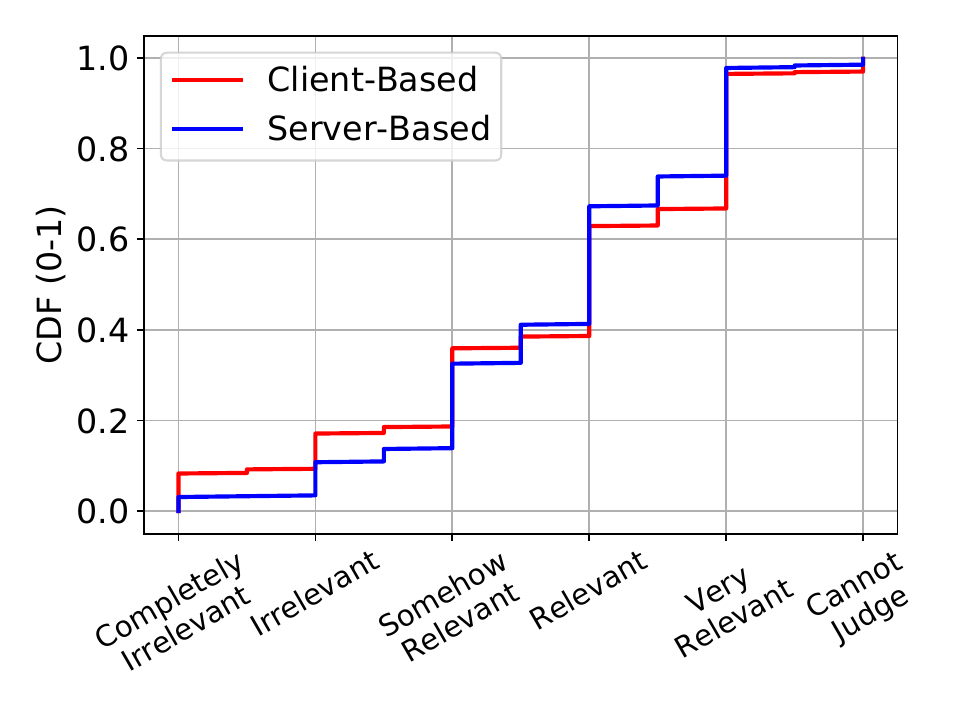}
    \caption{CDF of how relevant client-based and served-based annotations are with respect to original images extracted from webpages (\numimagesLarge images from \numpagesLarge webpages evaluated by \numpartLarge people each evaluating 10 random images).}
    \label{fig:scale}
\end{figure}

%%%%%%%%%%%%%%%%%%%%%%%%%%%%%%%
\section{On the Feasibility, Quality, and Performance of AI-Generated Webpages}
\label{sec:eval}
%%%%%%%%%%%%%%%%%%%%%%%%%%%%%%%
This section evaluates the applicability of generative AI for the Web, specifically focusing on stable diffusion for image generation~\cite{Ramesh_2022_DALLE2, Rombach_2022_CVPR, midjourney}. Our evaluation revolves around three main parts. First, we crawl \numpagesLarge webpages and quantify the \textit{feasibility} of the techniques discussed in the previous section for image annotation, which is paramount for the generation of the prompts used for image generation. We consider both client-based annotations, \ie a best effort to locate webpage text which should refer to an image, and server-based annotations, where we assume access to the original image as an approximation of a Web designer intent. Next, we evaluate the \textit{quality} of AI-generated webpages, or webpages where images are automatically generated via stable diffusion~\cite{Rombach_2022_CVPR}. We do this by performing several user studies involving about 900 participants provided by Prolific~\cite{prolific}. Finally, we benchmark the \textit{performance} of AI-generated webpages using the following web performance metrics metrics~\cite{web_performance}: SpeedIndex, Page Load Time, and bandwidth usage. We use \tool to perform all the experiments discussed in this section, leveraging   WebPageTest~\cite{webpagetest} and Chrome to automate the webpage loads and collect the browser telemetry data.

%%%%%%%%%%%%%%%%%%%%%%%%%%%%%%%%%%%%%%%%%%%%%%%%%%%%%%%
%\subsection{Feasibility of Images Annotation}
\subsection{Can Web Images Be Automatically Generated?}
\label{sec:res:feasibility}
%%%%%%%%%%%%%%%%%%%%%%%%%%%%%%%%%%%%%%%%%%%%%%%%%%%%%%%
We crawl (using Selenium) the top 500 webpages from Tranco's top one million list. About 25\% of these webpages fail to load;  for the remainder 75\% of webpages, the crawler achieves a 70\% success rate (260 webpages); failures are due to webpage formatting, \eg lack of \texttt{img} tags with clear \texttt{src} or \texttt{background-image} attributes. Next, we filter webpages hosting inappropriate or non-English content, given that stable diffusion currently operates only with English prompts. This results in \numpagesLarge webpages with a total of \numimagesLarge images for which some client-based annotation is produced. We finally run the \texttt{GenerativeImage2Text} transformer to derive the extra annotation needed by the  server-based scheme. This allowed us to generate a textual description of the original images.

Client-based annotation is a best effort attempt; as such, potential mistakes are possible. The output of the \\ \texttt{GenerativeImage2Text} transformer is instead highly accurate -- given their usage of BLIP image-captioning technique which is shown to outperform all state-of-the-art techniques~\cite{li2022blip} -- but generic. For example, an image of the Golden Gate bridge might generate a label such as ``a bridge with some clouds in the background''; conversely, the text in the webpage might describe more precisely which kind of bridge and its scenery (see Figure~\ref{fig:img_prompts} in the appendix for additional examples on the difference between both texts). 

We evaluate the correctness of the annotations using Natural Language Processing (NLP) of sentences similarity. We use the  \texttt{bert-base-nli-mean-tokens}~\cite{bert_base_tokens} model to generate text embeddings for both annotation schemes, each containing 768 values;  we then compute the cosine similarity~\cite{cosine} between both texts' embeddings. The results show that both the median and average similarities are about 50\% (with 75th percentile at 60\%, and 25th percentile at 40\%). This result is encouraging since it suggests some intersection between the two annotation schemes. Of course, perfect intersection is unlikely given that the output of the \texttt{GenerativeImage2Text} transformer is highly generic. However, systematic misbehavior of the client-based annotation scheme would result in much lower cosine similarity values.

%annotation extracted text per image in two different way: a) standard natural language processing (NLP) of sentences similarity, and b) user study using Prolific. For NLP, we compare the images annotation derived via client-based and server-based annotation. Comparing these two different text might not be ideal, given the fact that the web extracted text would not necessarily talk about the image but rather article behind the image, and the img2prompt text usually is a detailed description of the image without the context of the web article.

To further comment on the quality of automated web images annotations, we perform a user study. We extend \tool's Web interface (activated with parameter \texttt{?type=[scale, scale\_prompt]}, see Section~\ref{sec:meth:gui}) to show a single (original) image along some descriptive text and ask for how \textit{relevant} such text is to the image. Scores are in the range of ``completely irrelevant'' to ``very relevant''. We also allow study participants to respond ``cannot judge'', for the cases when participants are not knowledgeable of the image, \eg when the text refers to a celebrity they do not know. 

\begin{table}[t]
\small
\centering
\begin{tabular}{|l|c|c|c|c|c|} 
\hline
    &  \textbf{Images}  & \textbf{Images} & \textbf{Webpages} &     \textbf{Webpages} \\
    &  \textbf{Server}   & \textbf{Client} & \textbf{Server} & \textbf{Client} \\
    \hline\hline
    \textbf{Size} & 409 & 409 & 60 & 60 \\
    \hline
    \textbf{Participants} & 410 & 413 & 60 & 60 \\
    \hline
    \textbf{Responses} & 4,110 & 4,160 & 600 & 600 \\
    \hline
    \textbf{Age Range} & 19-63 & 19-63 & 20-40 &  18-52 \\
    \hline
    \textbf{Gender (M/F)}  & 239/171 & 197/216 & 20/10 & 20/10 \\
    \hline
    \textbf{\# Countries} & 29 & 26 & 9 & 10 \\
     
\hline
\end{tabular}
\caption{User study summary. \textit{Server} is short for server-based, and \textit{client} is short for client-based.}
\label{table:user}
\end{table}

Figure~\ref{fig:scale} shows the Cumulative Distribution Function (CDF) of the median score (10,000 scores from \numpartLarge people) for the \numimagesLarge images we have collected from \numpagesLarge websites, distinguishing between the client-based and server-based annotations shown on top of each image. The figure shows a success rate -- score higher than ``irrelevant'' --  of 80\% for the client-side approach, and close to 90\% for the server-side approach. This result has two important implications. First, not even the highly accurate \texttt{GenerativeImage2Text} is capable of being 100\% ``relevant'', according to our study participants. This is not unexpected, especially with no control on the input images which can assume a very abstract form (see Figure~\ref{fig:img_prompts} in the Appendix for sample images with their text description). It follows that the accuracy of the client-based approach is quite high, as it is only 10\% worse from the best attainable by a sophisticated AI.  

% \begin{table}[t]
% \centering
% \begin{tabular}{|l|c|c|c|c|} 
% \hline
%     \textbf{Type}   & \textbf{Size} & \textbf{Participants} & \textbf{Responses} \\
%     \hline\hline
%     Images & & & \\
%     Client-Based   & 409  & 411    & 4,110 \\ 
%     \hline
%     Images & & & \\
%     Server-Based   & 409  & 416    & 4,160 \\ 
%     \hline
%     Webpages & & & \\
%     Client-Based   & 25  & 25    & 250 \\ 
%     \hline
%     Webpages & & & \\
%     Server-Based   & 25  & 25    & 250 \\ 
% \hline
% \end{tabular}
% \caption{User study summary}
% \label{table:user}
% \end{table}

\begin{figure*}[t]
\centering
    \subfigure[CDF of median score per image comparison,  distinguishing between client and server-based image generation.]{\psfig{figure=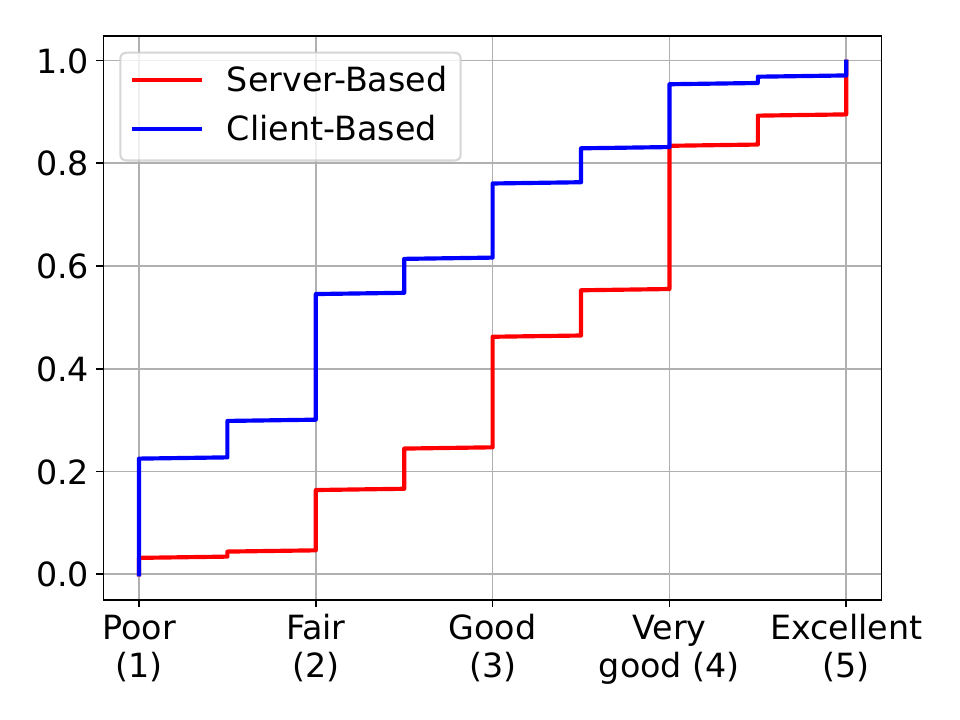, width=2.25in}\label{fig:eval:a}}
    \subfigure[Boxplot of median image scores as a function of their tags. Server-based image generation.  ]{\psfig{figure=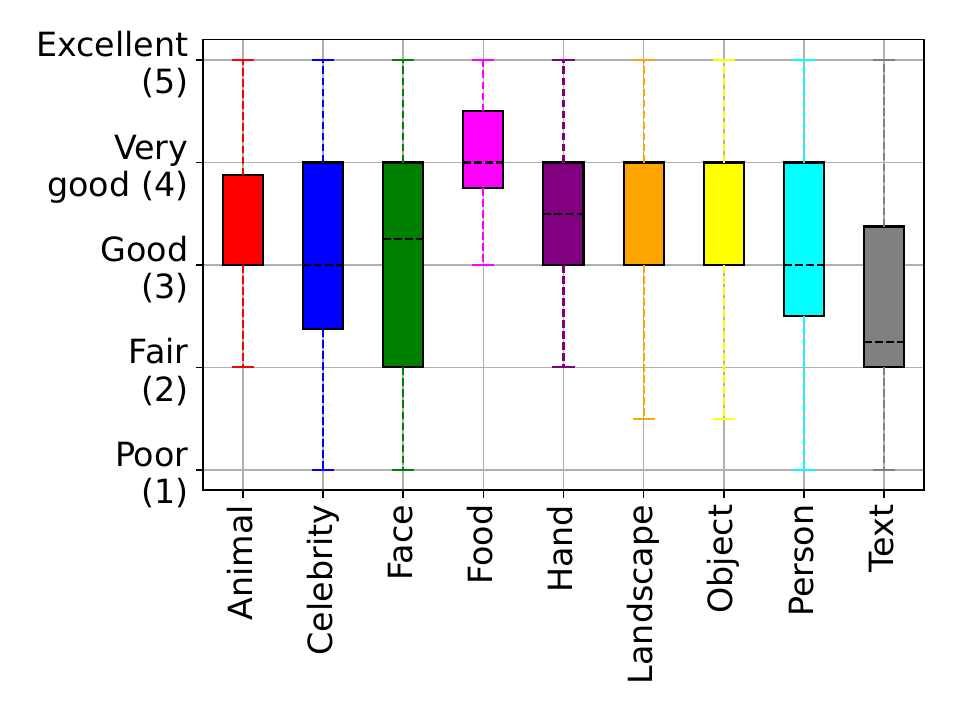, width=2.25in}\label{fig:eval:b}}
    \subfigure[CDF of median score per webpgage comparison, distinguishing between client and server-based image generation.]{\psfig{figure=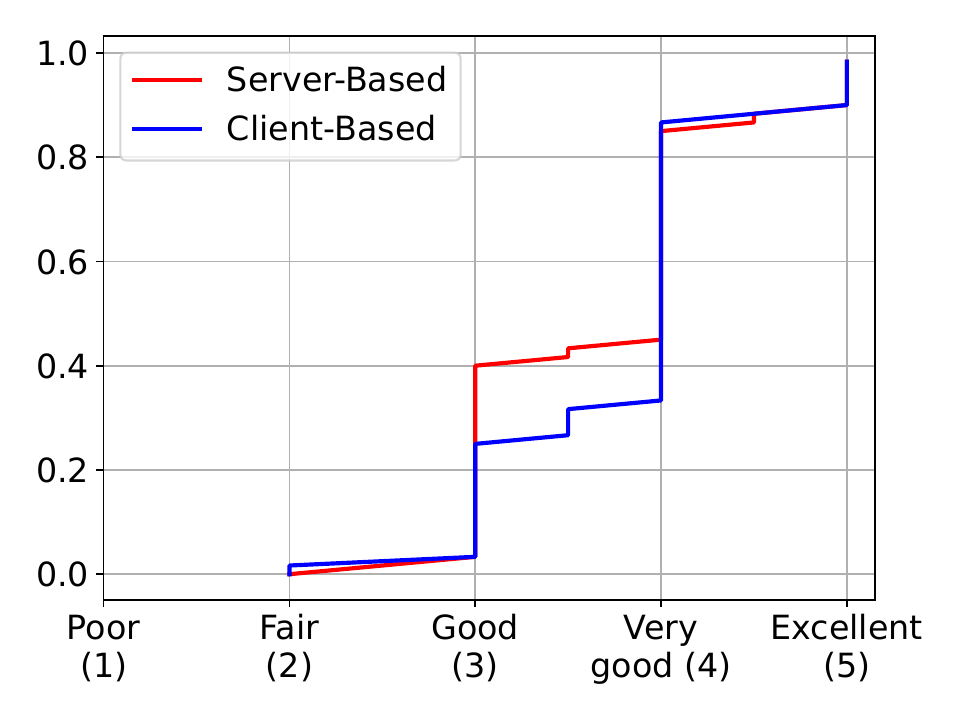, width=2.25in}\label{fig:eval:c}}
    %\vspace{-0.5in}
    \caption{Results from the user study. \textit{Client-based} refers to image generated on the client side, \ie only using contextual information like \texttt{alt text}.  \textit{Server-based} refers to image generated on the server side, \ie assuming availability of both contextual information and original image.}
   \label{fig:eval}
\end{figure*}

%MV - IMHO, these figures can go in the text, not much content provided.. 
% \begin{figure}[t]
%     \centering
%     \includegraphics[width=2.25in]{figs/SDXL_clipscore.pdf}
%     \caption{Boxplot of image clip scores as a function of their type: Original, Client-based, or Server-based images.}
%    \label{fig:eval:CLIP}
% \end{figure}

% \begin{figure}[t]
%     \centering
%     \includegraphics[width=2.25in]{figs/SDXL_gpu_inference_time.pdf}
%     \caption{StableDiffusion inference time as a function of the utilized GPU.}
%    \label{fig:eval:GPU}
% \end{figure}

%%%%%%%%%%%%%%%%%%%%%%%%%%%%%%%%%%%%%%%%%%%%%%%%%%%%%%%%
%\subsection{Quality of AI-Generated Webpages}
\subsection{What Is the Quality of AI-Generated Webpages?}
\label{sec:res:quality}
%%%%%%%%%%%%%%%%%%%%%%%%%%%%%%%%%%%%%%%%%%%%%%%%%%%%%%%%
This subsection investigates the quality of AI-generated webpages using user feedback collected on Prolific~\cite{prolific}. We first collect feedback per image composing a webpage, and then focus on the full webpages. Table~\ref{table:user} summarizes the user study. We collect 10 user feedback for the two versions (client-based and server-based) of the 409 images extracted from 60 websites randomly selected from the \numpagesLarge previously crawled. Next, we also collect 10 user feedback for two versions of each webpage, for a total of \tot user feedback. We ask Prolific to offer high quality participants which are fluent in English, so that they can understand the image descriptions. Among \numpart  study participants, the minimum \textit{approval rate}, \ie the past rate of approval of their work, we recorded was 94\%, with 80\% of the participants having an approval rate of 99-100\%. 

We start by investigating the quality of AI-generated Web images. Figure~\ref{fig:eval:a} shows the CDF of the median score (from 10 users) among  images, distinguishing between client-based and server-based. The figure shows that, regardless of the approach, the majority of scores are \textit{positive}, \ie ``fair'' or higher. Precisely, 95\% of the median scores are better or equal than ``fair'' for a server-based image generation; this number drops to 70\% in the client-based approach. A reduction is expected given the more challenging conditions (lack of access to the original image) in which the client-based approach operates.

Next we dig deeper into the collected scores with respect to the \textit{content} of the images. We manually tag each (original) image using the set of tags shown on the x-axis in Figure~\ref{fig:eval:b}; we limit to a maximum of three tags per image, by considering the three predominant features. For example, the image in Figure~\ref{fig:user_study} is labelled as ``object''. Note that we see the tags ``person'' and ``face'' which might seem redundant; however, these two tags are never used together, and their selection is based on whether a face is the main feature of an image. For example a closeup of a person would be labelled as ``face'', versus a group of people working is labelled as ``person''. 

Figure~\ref{fig:eval:b} shows one boxplot per image label, where each boxplot contains the median scores of all the images tagged with such label. We only show results for the server-based image generation scheme since it shows a similar trend as the client-based approach. If we focus on the median,  the figure shows that images containing ``food'', ``landscape'', and ``object'' achieve the best performance, with a median score of ``very good'' and rarely showing negative scores. Next, images labelled as ``hand'' or ``animal'' also perform well, with median comprised between ``good'' and ``very good''. It is worth noting that for all of the above categories even the 25th percentile of the boxplot still achieves a score equal to or above ``good''. When it comes to images related to humans (those labelled with  ``celebrity'', ``person'', or ``face''), we notice that their median scores are still equal to or above ``good'', however, their 25th percentile falls below that, even reaching a score of ``fair'' in the case of the ``face'' label. The only images that their median scores fall close to the ``fair'' score are those containing text. However, even in this category, the 25th percentile of the boxplot does not fall below the ``fair'' score. We conjecture that these low scores are due to fact that many of the text generated by stable diffusion are either not completely readable, or that it does not make sense  due to the inability of stable diffusion to re-produce accurate text. The opposite is instead responsible of the higher scores, \eg stable diffusion seems to be able to generate very convincing food images.  

\begin{figure*}[t]
\centering
    \subfigure[CDF of the delta for FCP, SI, and PLT (with two different GPUs).]{\psfig{figure=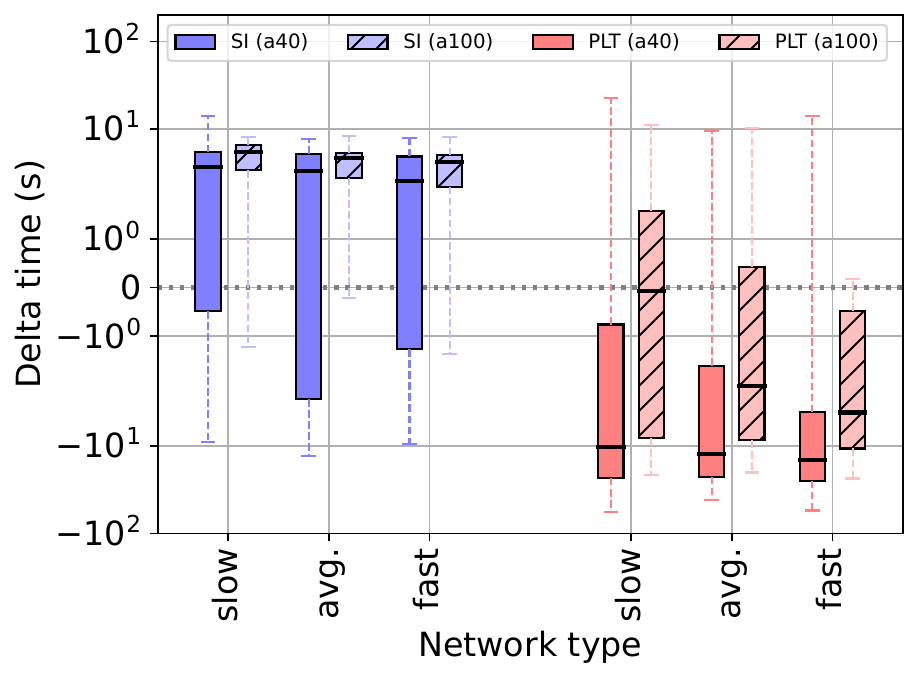, width=2.25in}\label{fig:eval2:a}}
    \subfigure[SD bandwidth savings in MB.]{\psfig{figure=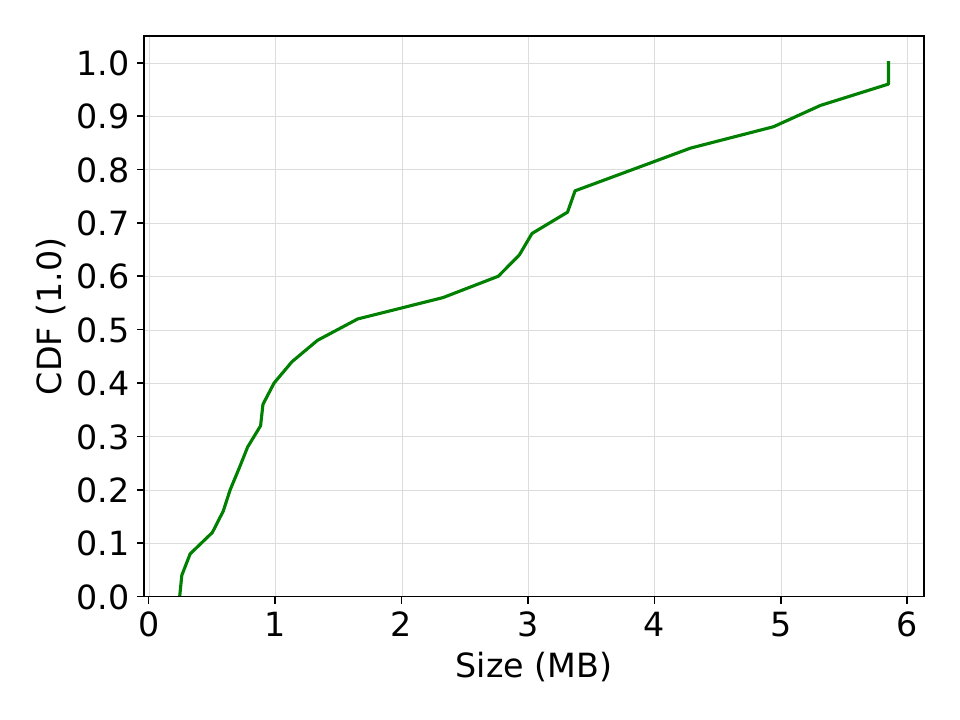, width=2.25in}\label{fig:eval2:b}}
    \subfigure[SD inference time in seconds across multiple GPU types.]{\psfig{figure=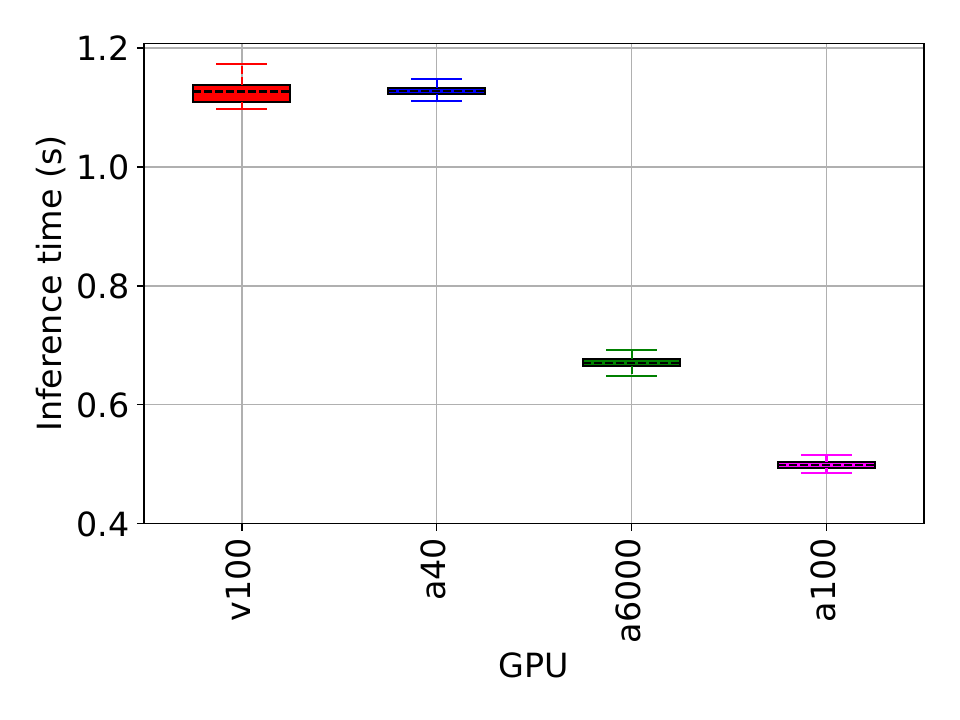, width=2.25in}\label{fig:eval2:c}}

    %\vspace{-0.5in}
    \caption{Reality check on running stable diffusion in the client for image generation.}
   \label{fig:eval2}
\end{figure*}

Finally, we comment on the quality of AI-generated webpages. Figure~\ref{fig:eval:c}  shows the CDF of the median score (from 10 users) among  webpages, distinguishing between the client-based and server-based approach. Compared to Figure~\ref{fig:eval:a}, the overall score has improved, with no scores being ``poor'', and only a few being ``fair''. Indeed, 55-68\% of the scores are either ``very good'' or ``excellent''. The reason for the score increase compared to the previous image to image comparison is that webpages are more complicated, composed by a collection of text/images. While for the images study participants can focus on low level details, such as potential hands deformity, this is less likely on a webpage where eyeballs tend to move quickly, potentially ignoring some images. The results of the AI-generated webpages show similar curves for both the client-based and the server-based, apart from the gap in the region between ``good'' and ``very good''. In this region, about 10\% of the images have scored higher in the case of client-based (``very good''), and lower in the case of server-based (``good''). We conjecture that the quality difference of these images is so subtle, making it a very difficult and subjective decision on which of these two scores, ``good'' vs.\ ``very good'', should these images be assigned. 

%%%%%%%%%%%%%%%%%%%%%%%%%%%%%%%%%%%%%%%%%%%%%%%%%%%%%%%%%
%\subsection{Client and Webpage Performance}
\subsection{Can Image Generation Run in the Client?} 
%%%%%%%%%%%%%%%%%%%%%%%%%%%%%%%%%%%%%%%%%%%%%%%%%%%%%%%%%%%
We conclude our analysis by performing a reality check on running stable diffusion directly in the client. This technology offers several benefits, such as additional privacy and rectification of broken web pages, but comes at the cost of expensive hardware and/or potential webopage slowdowns.

For these experiments, we assume a powerful desktop machine similar to our stable diffusion server, \ie mounting powerful GPUs like the A40 or the A100, since a mobile phone's hardware is far from being capable of running such complex models. For example, as of June 2023, it took an iPhone-13 about 15.6 seconds to generate a single image using stable diffusion leveraging apple's new core ML framework~\cite{iphone_SD}\footnote{This is viewed as a significant improvement compared to the 23 seconds generation time back in December 2022.}. For this reason, we emulate typical speeds of home connections~\cite{etisalat_internet} and ignore slower speeds like 3G. Network emulation is realized  using \texttt{TC} (see Figure~\ref{fig:setup}) instructed via the following connectivity profiles: \textit{slow} (symmetric 20~Mbps and 100~ms RTT), \textit{average} (symmetric 50~Mbps and 50ms RTT), and \textit{fast} (symmetric 100~Mbps and 20ms RTT). We use Chrome to load the previous 60 webpages (repeating each 5 times), both in their original form and considering AI-generated images. We use textual prompts derived with the server-based scheme, although we did not measure any difference in the duration of image generation using the two approaches. 

%FCP measures the time from when the page starts loading to when any part of the page's content is rendered on the screen. 
%SI: -2.3255
%PLT: -10.5125
%99 VC -5.0

\vspace{0.05in}
\noindent
\textbf{Web Performance:} We focus on two popular Web performance metrics: SpeedIndex~ (SI), and Page Load Time (PLT) ~\cite{web_performance}. SI measures how quickly the visible content of a webpage is displayed during a page load. Conversely, PLT measures the total time it takes the browser to download and visualize the entire webpage, including content located below the fold. Figure~\ref{fig:eval2:a} shows the CDF of the \textit{delta} for each metric, derived as the difference between the metric computed for the original webpage, and the metric computed for the webpage when images are AI-generated. It follows that a negative value indicates a slowdown, and a positive value indicates a speedup. Each delta value in the plot is the median out of 5 runs. 

The figure shows that powerful GPUs (A40 and A100) can compete with actual image downloads when considering SI, \eg 63\% (A40) and 87\% (A100) of the websites benefit from a speedup even in presence of a very fast network. This happens because, on average, only 2-5 images are loaded above the fold, and the transport might need some time to converge to the available bandwidth, \eg due to connection establishment and window increase. Furthermore, certain images positioned above the fold have substantial file sizes. However, when we consider \texttt{PageLoadTime}, even on a very slow network almost no webpage loads faster when images are locally generated. Whether this might change in the future is a question of how hardware and AI models would evolve, in contrast with network performance. Regardless, this result also suggests that locally generating images is already viable for applications operating on a handful of images, such as privacy-preserving advertising and webpage repairing -- granted that powerful GPUs are adopted.

\vspace{0.05in}
\noindent
\textbf{Resource Usage:} Figure~\ref{fig:eval2:b} quantifies the bandwidth savings (MB) due to not transferring Web images which are instead locally generated. Overall, the majority of webpages we tested enjoy multiple MBs of savings (median of 1.3MB), and even more than 5MB for the 10\% heaviest webpages. These savings are marginal for devices on WiFi, the likely connectivity scenario for the powerful devices we are assuming. However, these savings can be significant for the content providers. For example, let's consider a popular newspaper with 1 million daily views; a median saving of 1.3~MB per view would account for daily savings of 1.3~TB. 

Finally, Figure~\ref{fig:eval2:c} shows the boxplot of the stable diffusion image generation time (i.e., inference time), measured when generating our 409 images with four different popular GPUs. We can see that in the case of V100 and a40 the median inference time is about 1.1 seconds. However, with more powerful GPUs such as the a100, the median inference time drops down to about 500ms. It is also worth noting that the boxplots are extremely narrow, suggesting that the inference time does not vary across multiple textual prompts.

%\input ./sections/05_discussion
%%%%%%%%%%%%%%%%%%%%%%%%%%%%%%%%%%%%
\section{Conclusions and Future Work}
\label{sec:concl}
%%%%%%%%%%%%%%%%%%%%%%%%%%%%%%%%%%%%
Over the past year, we have observed a significant increase in the adoption of generative Artificial Intelligence (AI) across a range of fields, including medicine and media production. Similarly, Web developers and browser vendors are turning to generative AI for various applications like multimedia asset creation, webpage acceleration, and privacy enhancements. In this paper, we have delved into the role of generative AI in the specific context of web image generation and consumption. To accomplish this, we have introduced \tool, a tool that facilitates real-time experiments using unaltered web browsers, thus enabling an in-depth assessment of the quality and effectiveness of AI-generated images within target webpages. Moreover, by seamlessly integrating with crowdsourcing platforms such as Prolific, \tool offers a valuable avenue for collecting feedback and insights from a diverse global audience. Our results indicate that generative AI can already create relevant and high-quality web images without the need for manual prompts from web designers, relying on contextual information found within webpages. Conversely, in-browser image generation remains a challenging task, as it requires highly capable GPUs like the A40 and A100 to partially match traditional image downloads. Nevertheless, this approach holds promise for small scale image generation tasks, such as repairing broken webpages or serving privacy-preserving media.

This paper opens several interesting avenues for future work. First, stable diffusion often produces artefacts when generating images of faces and hands even with a large number of iterations (> 100, increasing the image generation time to above 15 seconds), especially when there are more than one subject in the produced image. While stable diffusion is constantly improving with this respect, one could consider extending \tool to other text-to-image generation models. For example, the GFPGAN model is a popular method that addresses the issue of distorted faces. This model helps restore faces in samples and improves the overall realistic quality of the generated image~\cite{gfpgan}.

Next, this paper currently only focuses on image generation. While images are a fundamental component of webpages, and account for a large fraction of the page size, there are other elements like HTML, CSS, and JavaScript which can be (partially) automatically generated. Such generation is however more challenging as, differently from images, it can impact Web compatibility, \ie breaks the functioning of webpages. Nevertheless, with the constant improvement of generative AI, it is possible to envision how these tools can be used in broader sense for the Web, and we thus plan to extend \tool to support them. 

\section{Ethics}
An institutional review board (IRB) approval is granted to conduct the study, and the authors who conducted the study are CITI certified.

%Bibliography
%\bibliographystyle{abbrv} 
\bibliographystyle{abbrv_compact} 
\bibliography{biblio}

\vspace{-5pt}
% appendix
% \clearpage

\appendix
%%%%%%%%%%%%%%%%%%%%%%%%%%%%%%%%%%%%%%%%%%%%%
\section{Appendix}
\label{sec:appendix}
%%%%%%%%%%%%%%%%%%%%%%%%%%%%%%%%%%%%%%%%%%%
In this appendix we show few images that received a low score during the user study associated with the feasibility evaluation (Section~\ref{sec:res:feasibility}). In addition, we also show three webpages in three forms: original, AI-generated -- using both the client-based and server-based annotation (Section~\ref{sec:meth:anno}). 

\begin{figure*}[hbt]
\centering
    \subfigure[A golden statue of a person holding a sword; Akriti brass art wares antique brass metal lord ganesha on fan wall hanging for entrance door; living room; decorative]{\psfig{figure=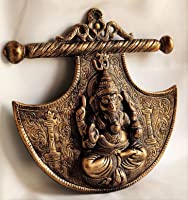, height=1.49in}}\hfill
    \subfigure[A colorful tree with a man and a dog on it; Mozilla puts people before profit; creating products; technologies and programs that make the internet healthier for everyone. learn more about us more power to you.]{\psfig{figure=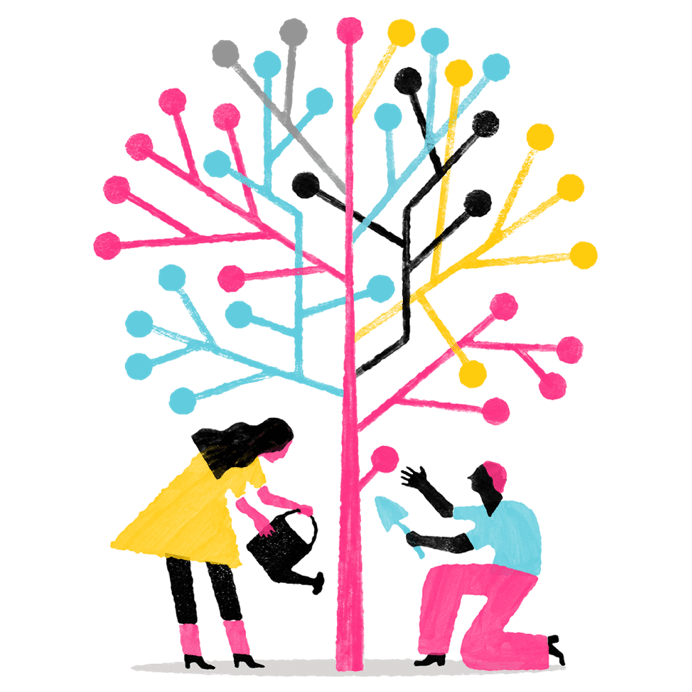, height=1.49in}}\hfill
    \subfigure[A hand is holding a pencil in a drawing style; An effective landing page needs a layout; a way to implement it; clear; concise copy; and ongoing evaluation—all with the goal of driving visitors to your cta. how to make a landing page]{\psfig{figure=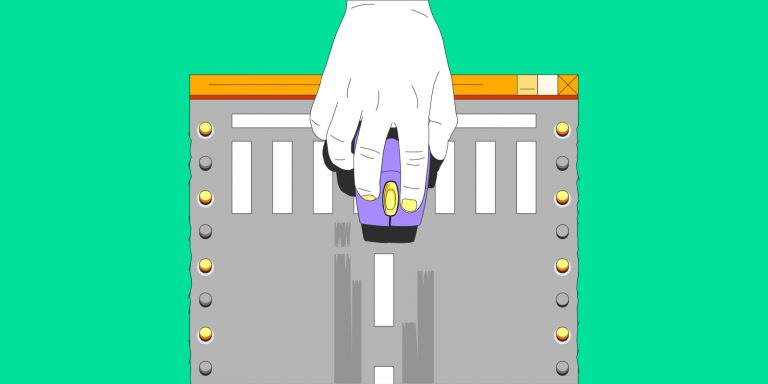, height=1.49in}}\hfill
    \subfigure[A castle made out of colored plastic blocks; Picassotiles 60 piece set 60pcs magnet building tiles clear magnetic 3d building blocks construction playboards]{\psfig{figure=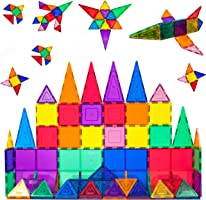, height=1.49in}}\hfill
    \subfigure[A man with a stack of cassettes on his head; By open mike eagle past and present blur on the new lp from the celebrated mc. read more a tape called component system with the auto reverse]{\psfig{figure=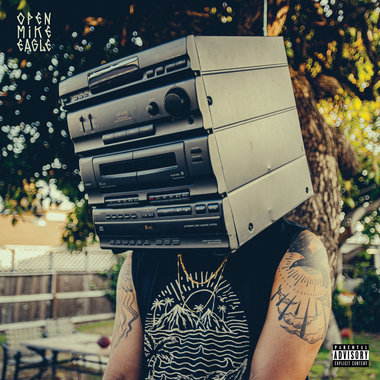, height=1.49in}}\hfill
    \subfigure[The front of a building with columns and a clock; Nih is the nations medical research agency; supporting scientific studies that turn discovery into health. nih-at-a-glance-whoweare.jpg who we are who we are]{\psfig{figure=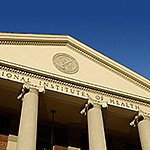, height=1.49in}}\hfill
    \subfigure[A woman sitting at a table using a laptop computer; Visit cisco hybrid work index to understand the inclusive collaboration experiences driving hybrid work. the future of work is hybrid]{\psfig{figure=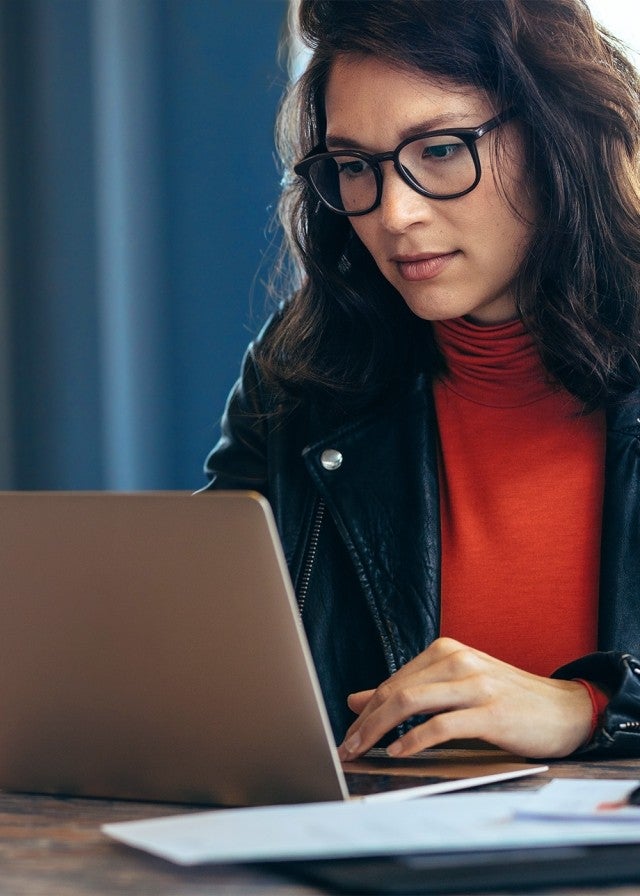, height=1.49in}}
    \subfigure[A couple of pictures of a girl and a boy; Your files and memories stay safe and secure in the cloud; with 5 gb for free and 1 tb+ if you go premium store with confidence visual renderings demonstrating some of the content types that can be safely stored in office]{\psfig{figure=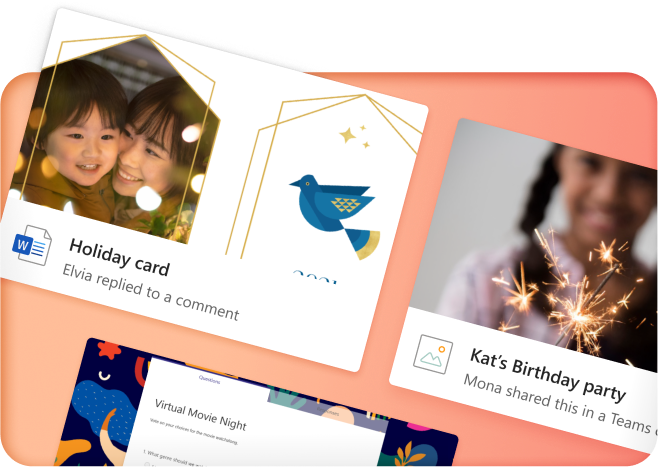, height=1.49in}}\hfill
    \subfigure[A woman standing in front of a large truck; Lady of the gobi;a rare woman among thousands of coal truck drivers fights for survival on the hazardous mining road from mongolia to china;watch now;25.13 coal truck driver; maikhuu; in the film lady of the gobi directed by khoroldorj choijoovanchig]{\psfig{figure=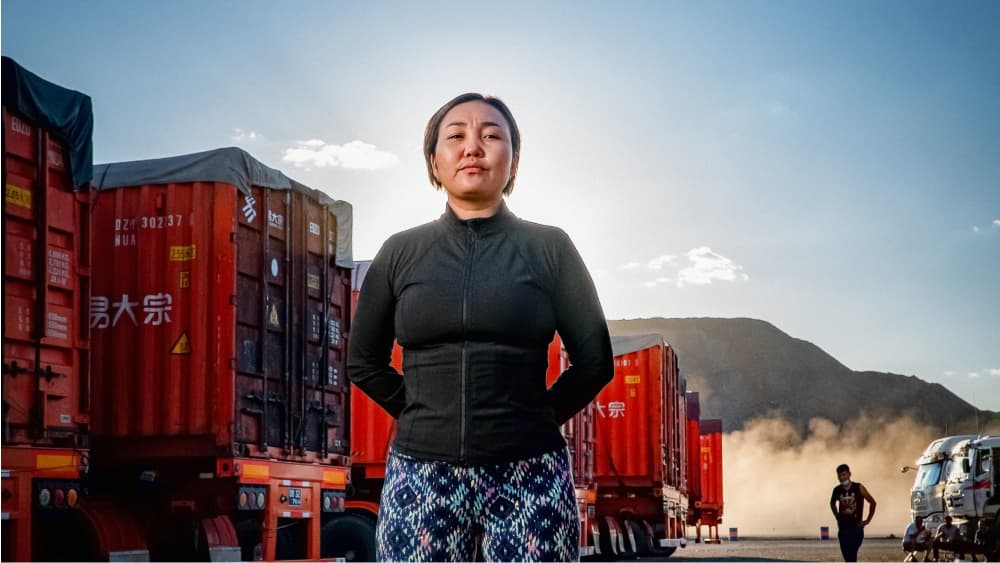, height=1.49in}}\hfill
    \subfigure[A black and white photo with the words all about it; All about the white lotus season 2]{\psfig{figure=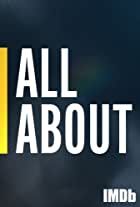, height=1.49in}}\hfill
    \caption{Several samples of Web images with their Image2Text (first sentence before the semicolon) and extracted text descriptions (text following the first semicolon) that were given a low score (``irrelevant'' or ``very irrelevant'') by our study participants.}
   \label{fig:img_prompts}
\end{figure*}

\begin{figure*}[hbt]
\centering
    \subfigure[Original]{\psfig{figure=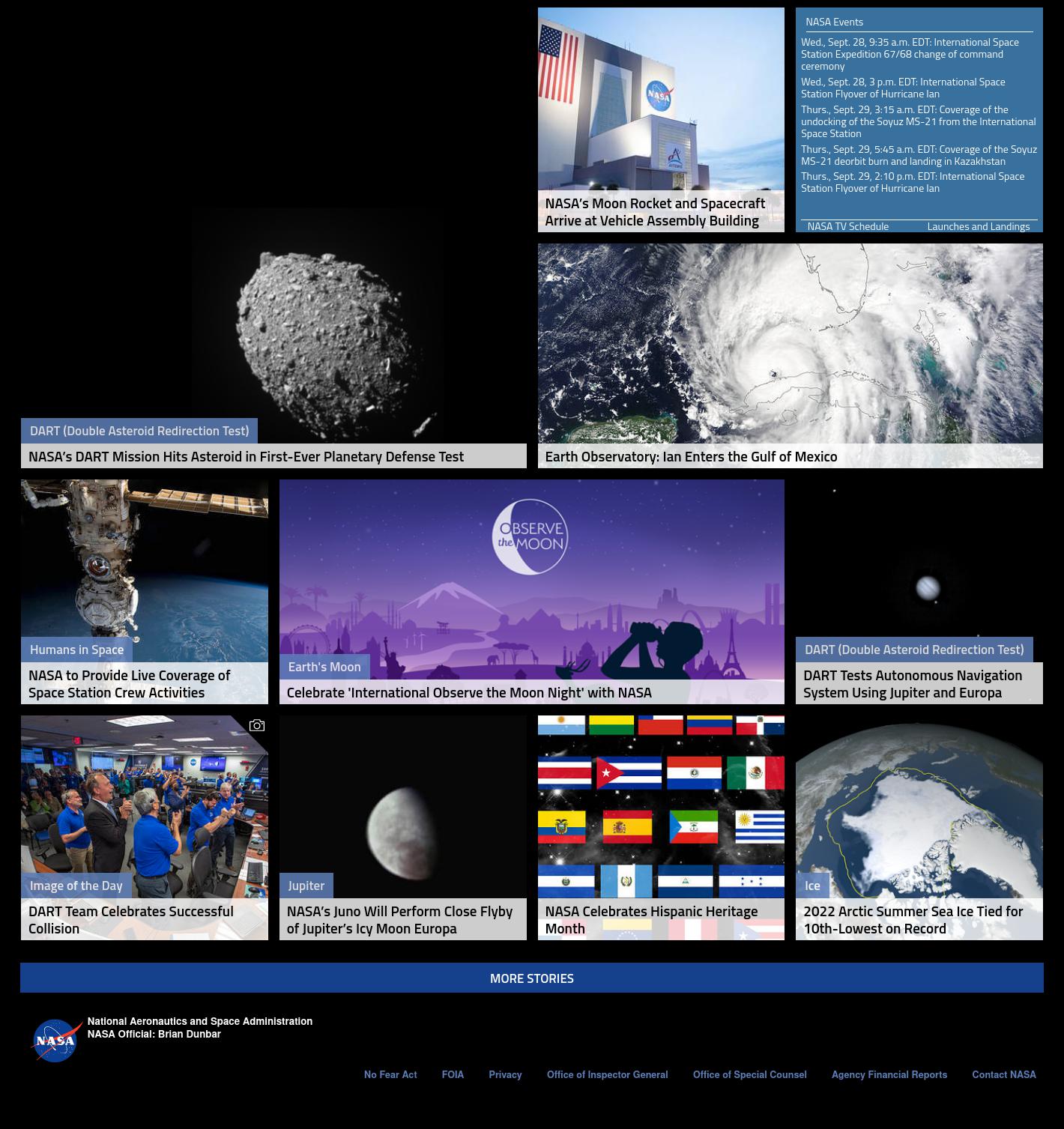, width=2.2in}}\hfill
    \subfigure[Client-based]{\psfig{figure=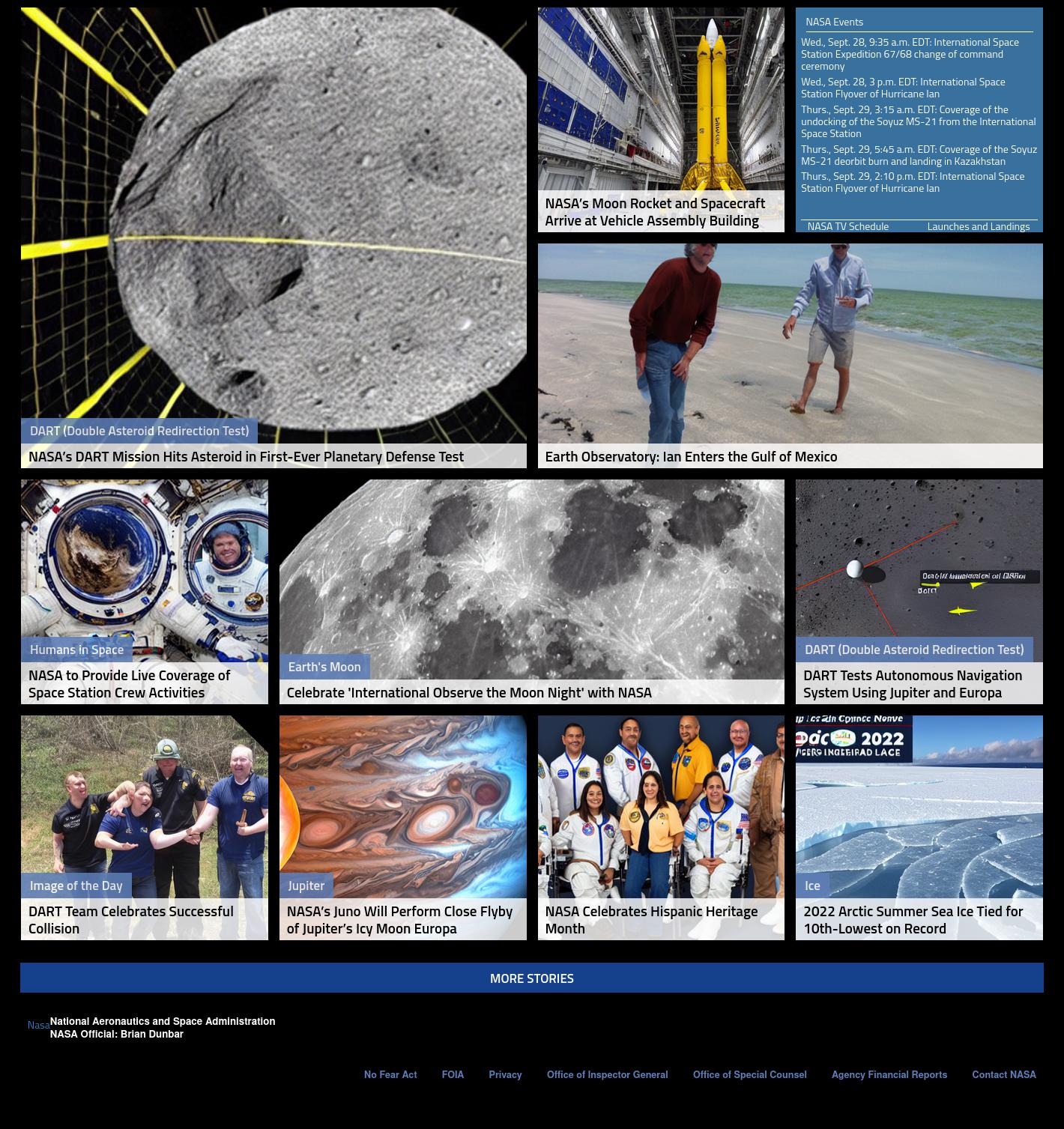, width=2.2in}}\hfill
    \subfigure[Server-based]{\psfig{figure=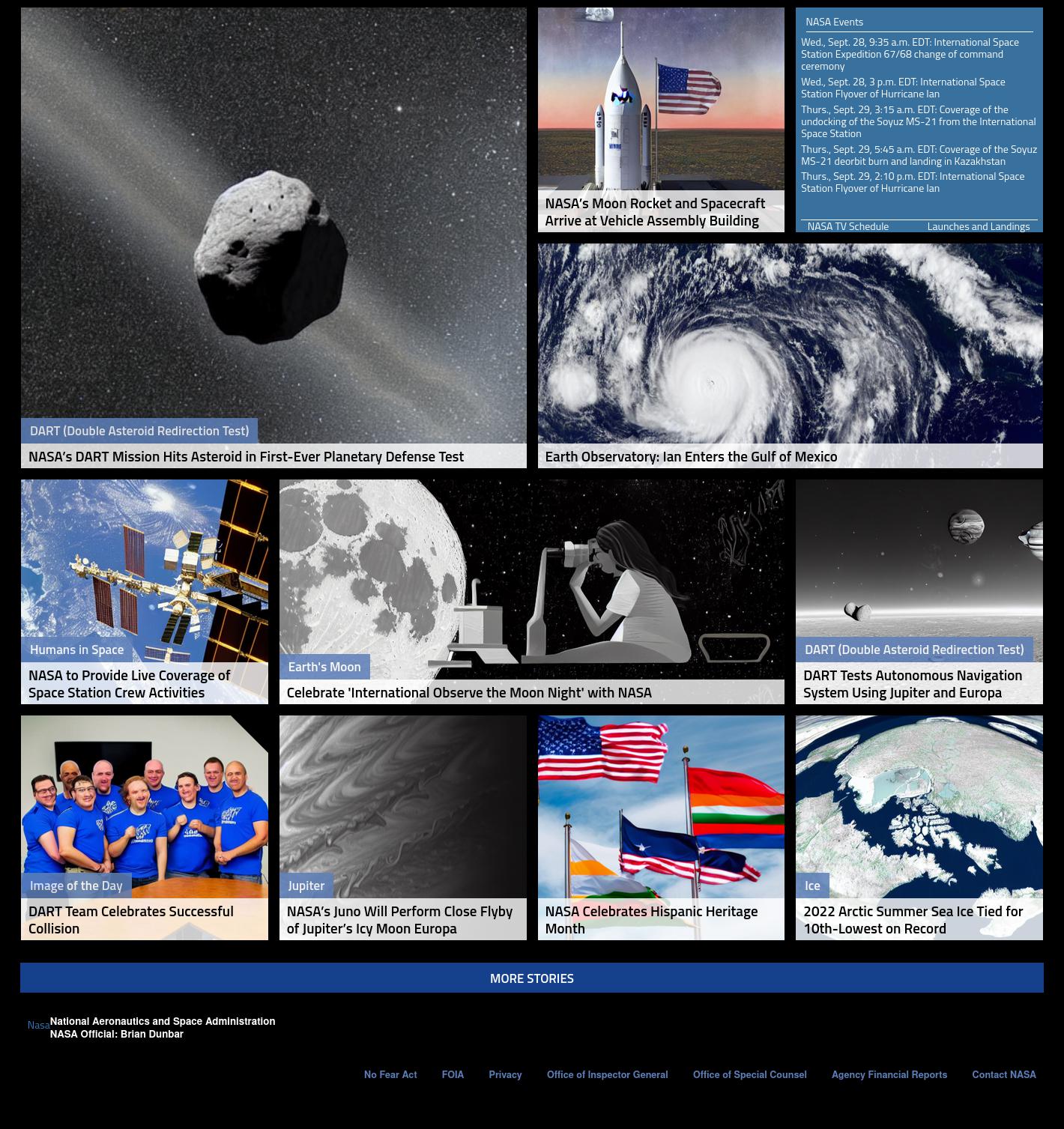, width=2.2in}}\hfill
    \caption{Comparing the www.nasa.gov webpage with webpages where images are automatically generated using stable diffusion and both the client-based and server-based approach.}
   \label{fig:nasa}
\end{figure*}

\begin{figure*}[hbt]
\centering
    \subfigure[Original]{\psfig{figure=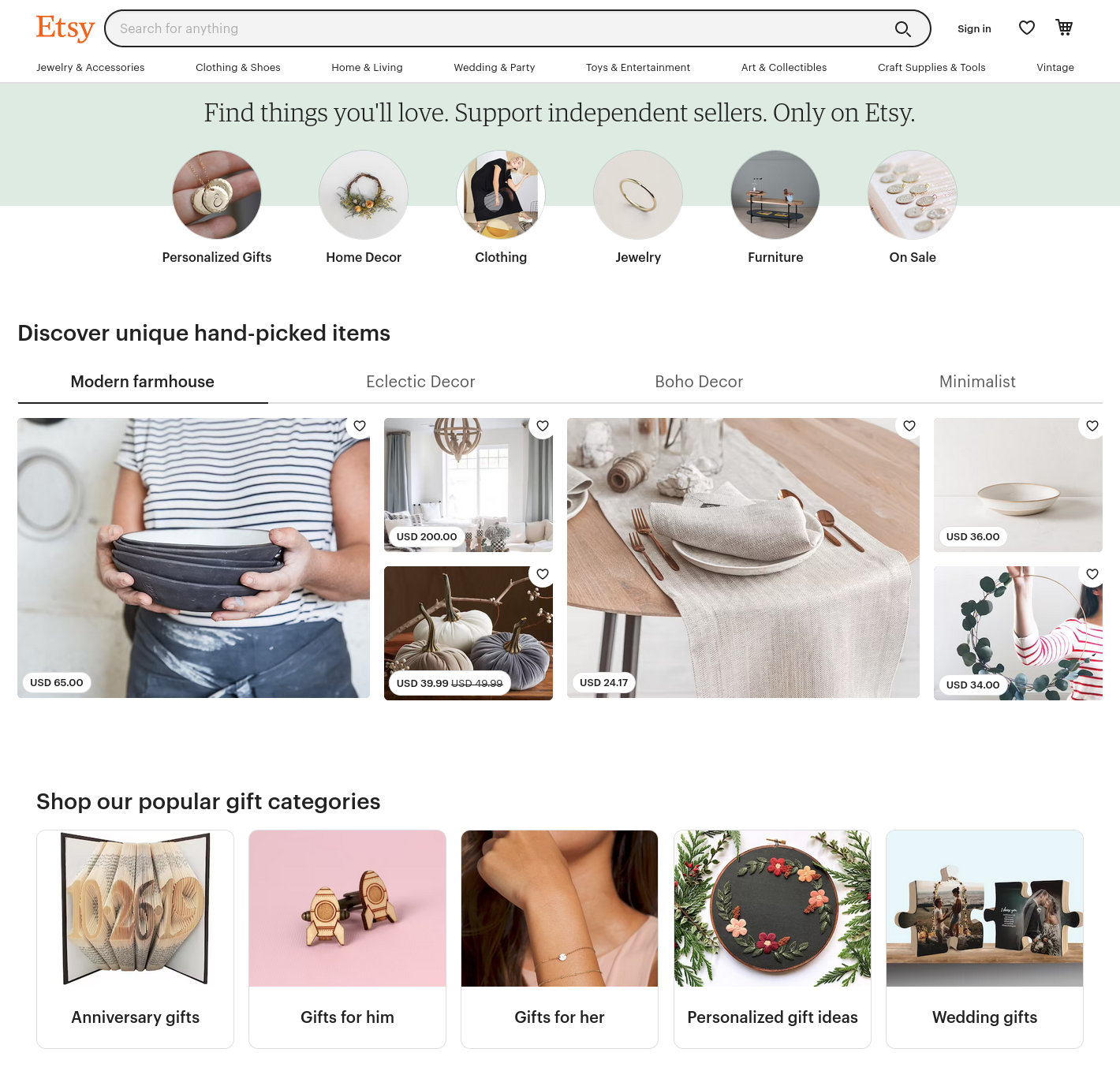, width=2.2in}}\hfill
    \subfigure[Client-based]{\psfig{figure=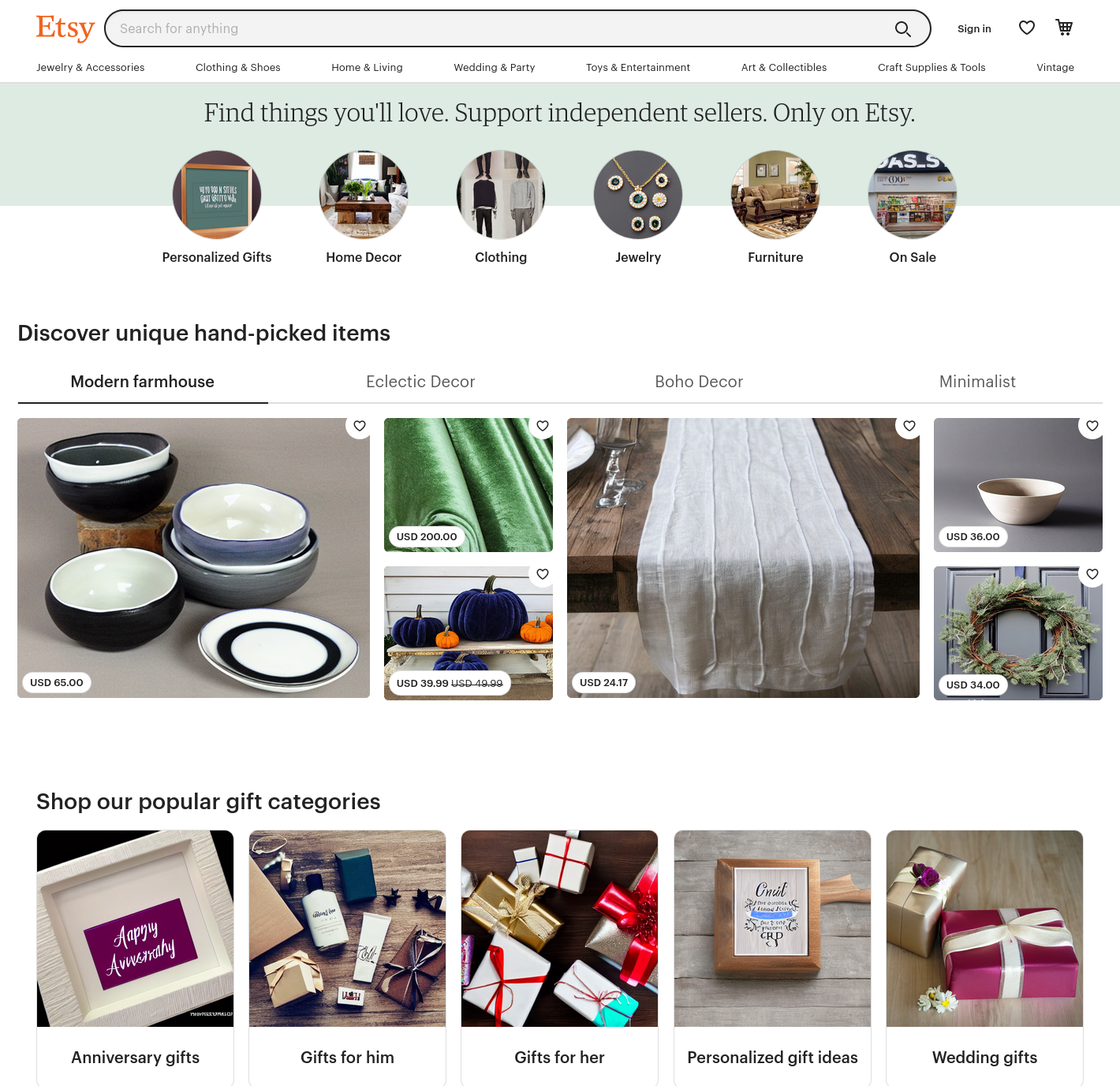, width=2.2in}}\hfill
    \subfigure[Server-based]{\psfig{figure=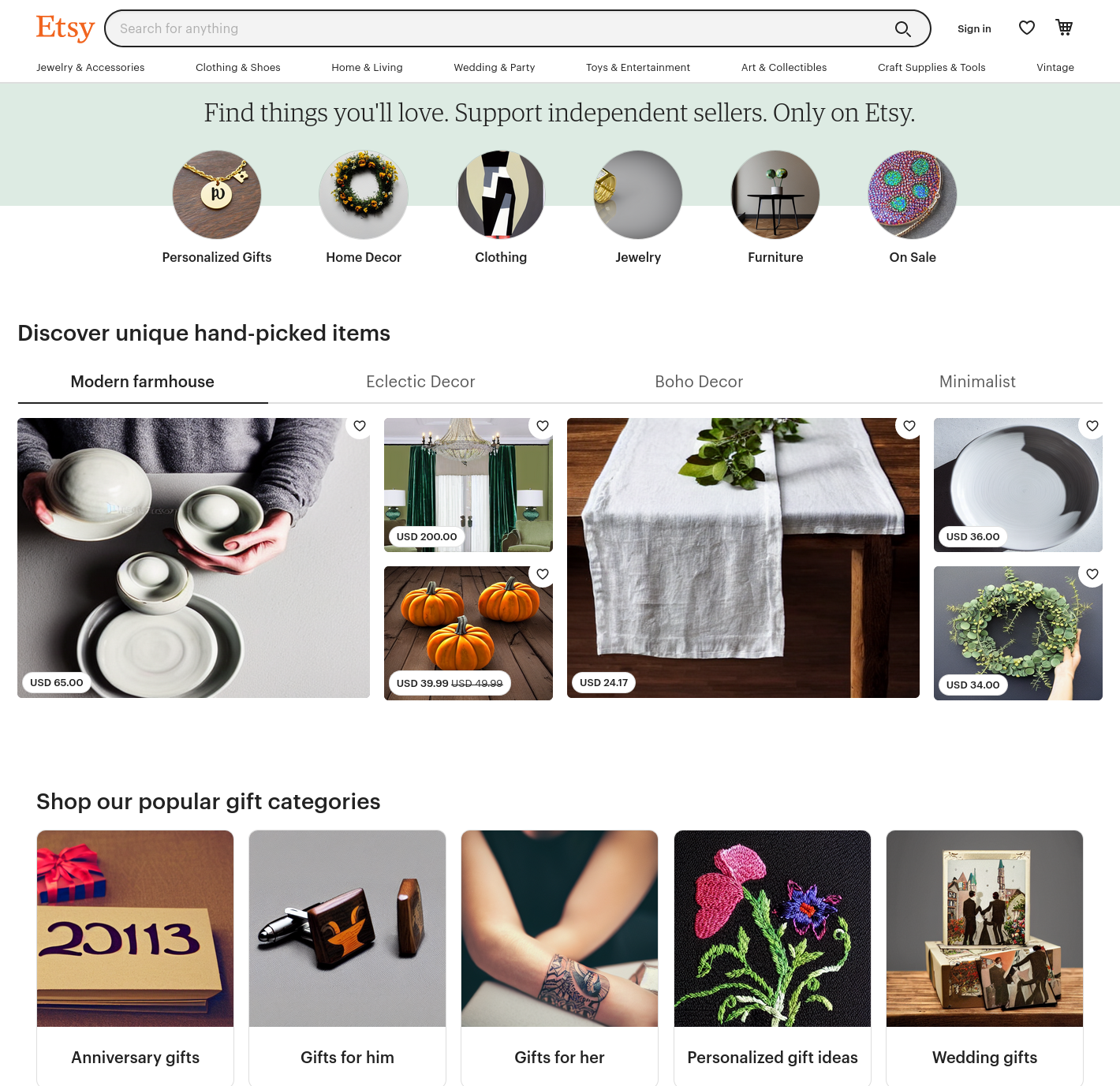, width=2.2in}}\hfill
    \caption{Comparing the www.etsy.com webpagewith webpages where images are automatically generated using stable diffusion and both the client-based and server-based approach.}
   \label{fig:etsy}
\end{figure*}

\begin{figure*}[hbt]
\centering
    \subfigure[Original]{\psfig{figure=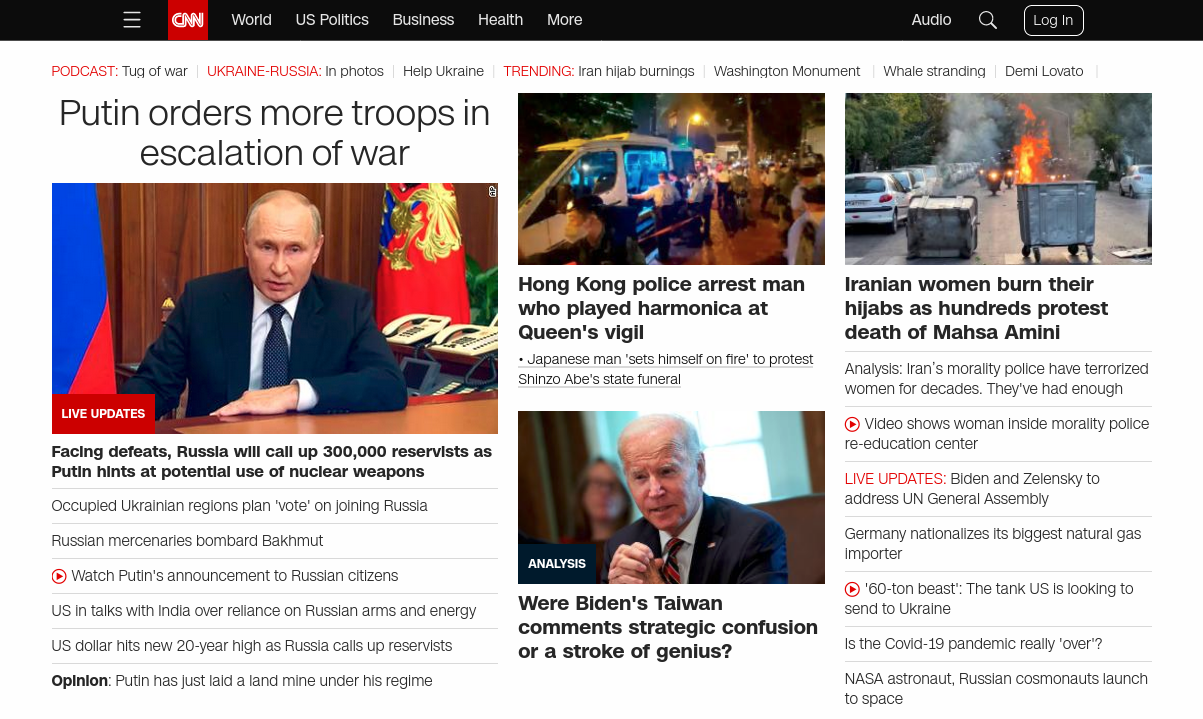, width=2.3in}}\hfill
    \subfigure[Client-based]{\psfig{figure=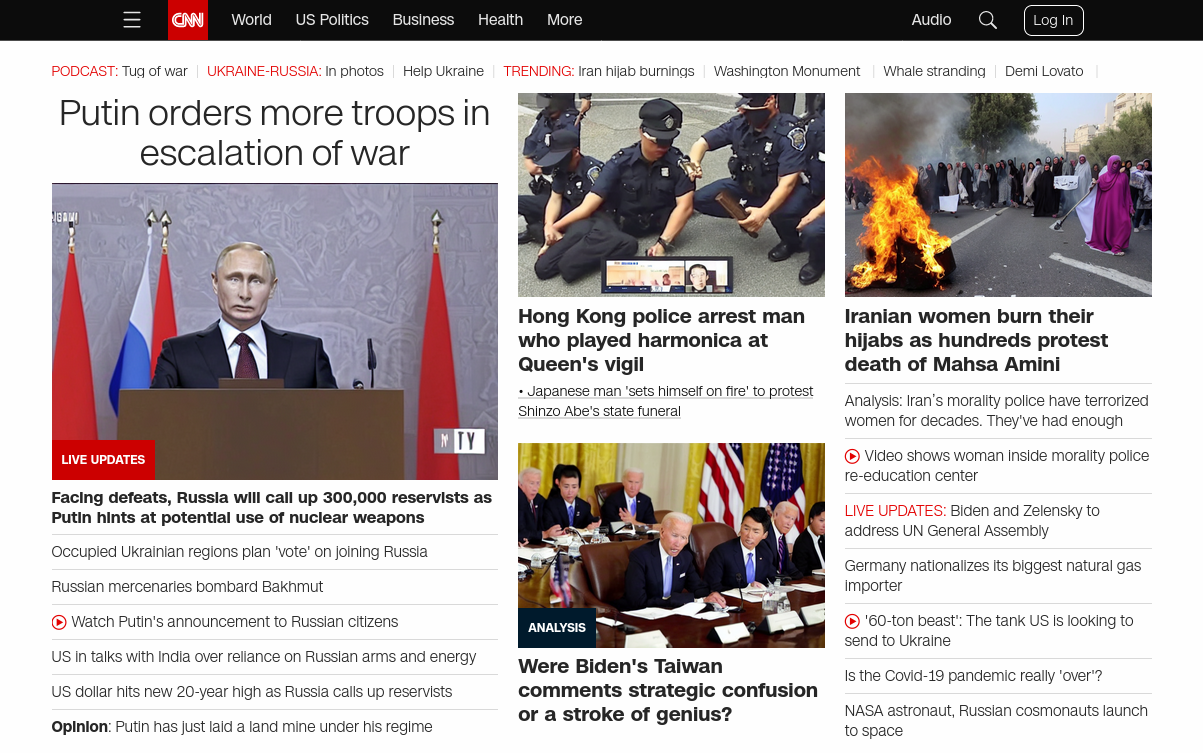, width=2.2in}}\hfill
    \subfigure[Server-based]{\psfig{figure=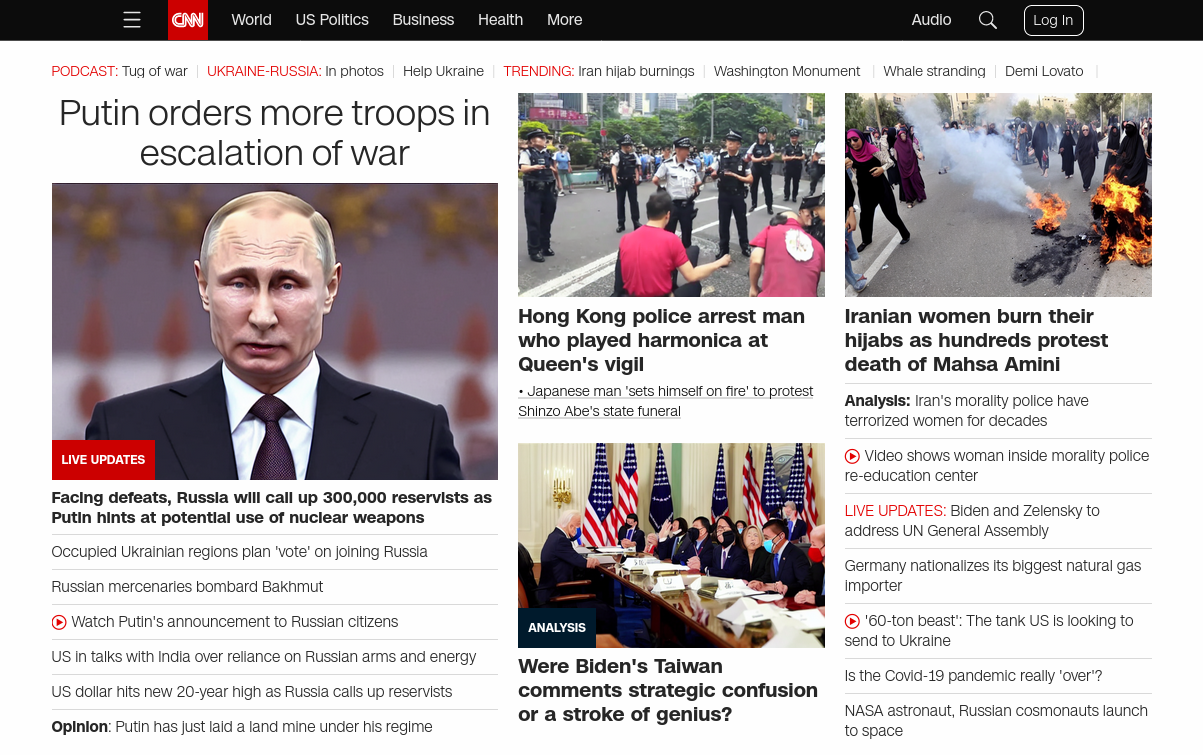, width=2.2in}}\hfill
    \caption{Comparing the www.cnn.com webpage with webpages where images are automatically generated using stable diffusion and both the client-based and server-based approach.}
   \label{fig:cnn}
\end{figure*}

\end{document}